\pdfoutput=1
\documentclass[11pt]{article}
\usepackage[final]{acl}
\usepackage{times}
\usepackage{latexsym}
\usepackage[T1]{fontenc}
\usepackage[utf8]{inputenc}
\usepackage{microtype}
\usepackage{inconsolata}
\usepackage{graphicx}
\usepackage{amssymb}
\usepackage{pifont}
\usepackage{colortbl}
\usepackage{array}
\usepackage{booktabs} 
\usepackage{ulem}
\usepackage{multirow}
\usepackage{amsmath}
\usepackage{graphicx} 
\usepackage{algorithm}
\usepackage{algpseudocode}
\usepackage{algorithmicx}
\usepackage[most]{tcolorbox}
\usepackage{tabularx}
\tcbuselibrary{breakable}
\usepackage{lipsum}
\usepackage{diagbox}
\usepackage{subfigure}
\usepackage{cuted}
\usepackage{multicol}
\usepackage{enumitem}
\usepackage{soul}
\usepackage[nottoc]{tocbibind}
\makeatletter
\newcommand\listofappendices{
    \section*{List of Appendices}
    \@starttoc{loa}
}
\makeatother
\newcommand{\bestbox}[1]{\setlength{\fboxsep}{1pt}\colorbox[rgb]{0.8,0.9,1}{#1}}
\newcommand{\bestusftbox}[1]{\setlength{\fboxsep}{1pt}\colorbox[rgb]{0.8,1,0.8}{#1}}
\newcommand{\worstbox}[1]{\setlength{\fboxsep}{1pt}\colorbox[rgb]{1,0.8,0.8}{#1}}

\definecolor{morandiPink}{RGB}{242,218,218}
\definecolor{morandiGreen}{RGB}{220,235,226}
\definecolor{morandiBlue}{RGB}{215,222,235}
\definecolor{placeholdercolor}{RGB}{106, 143, 163}
\definecolor{darkred}{RGB}{178,34,34}      
\definecolor{lightblue}{RGB}{51,102,187}     
\newcommand{\phcolor}[1]{\textcolor{placeholdercolor}{#1}}
\newcommand{\BLUE}[1]{{\textcolor{lightblue}{#1}}}
\newcommand{\RED}[1]{{\textcolor{darkred}{#1}}}

\newcommand{\qwen}{Qwen2.5-7B-Instruct}
\newcommand{\llamas}{Llama-3.1-8B-Instruct}

\newcommand{\llamal}{Llama-3.1-70B-Instruct}

\newcommand{\gptf}{GPT-4o}
\newcommand{\gptomini}{GPT-o1 mini}
\newcommand{\claude}{Claude-3.5-Sonnet}

\newcommand{\basic}{\textbf{DirectResponse}}
\newcommand{\cotA}{\textbf{FullAttributes}}
\newcommand{\cotAs}{\textbf{TaskRelated}}
\newcommand{\cotAAs}{\textbf{AttributeFilter}}
\newcommand{\cotTsAs}{\textbf{TypeGuided}}

\newcommand{\ipdialogfontPal}{\fontfamily{pnc}\selectfont}
\newcommand{\ipdialog}{{\ipdialogfontPal IP-Dialog}}

\title{\ipdialog~\includegraphics[width=0.6cm]{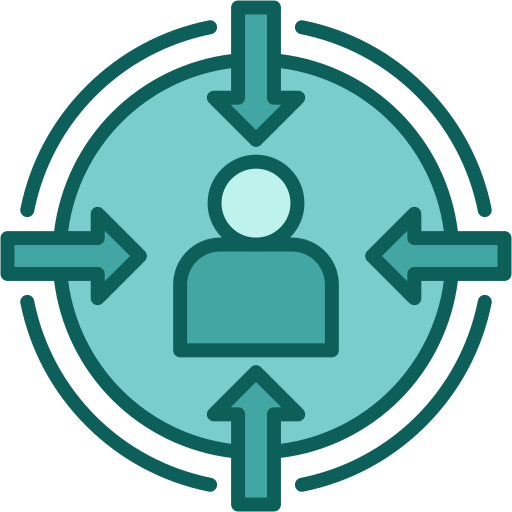}: Evaluating Implicit Personalization in Dialogue Systems with Synthetic Data}

\author{Bo Peng$^{1,2,3}$\footnotemark[1], Zhiheng Wang$^{1,2}$\footnotemark[1], Heyang Gong$^{4}$, Chaochao Lu$^{1,3}$\footnotemark[2] \\
  $^{1}$Shanghai Artificial Intelligence Laboratory,~~~$^{2}$Shanghai Jiao Tong University \\
  $^{3}$Shanghai Innovation Institute~~~$^{4}$Sicore Ladder Tech Co. Ltd.\\
  \texttt{peng\_bo2019@sjtu.edu.cn}, \texttt{wangzhiheng@pjlab.org.cn},\\
  \texttt{zj3712@gmail.com},
  \texttt{luchaochao@pjlab.org.cn}
  }

\begin{document}
\maketitle
\begin{abstract}
  In modern dialogue systems, the ability to implicitly infer user backgrounds from conversations and leverage this information for personalized assistance is crucial. However, 
  the scarcity of high-quality data remains a fundamental challenge to evaluating and improving this capability. Traditional dataset construction methods are labor-intensive, resource-demanding, and raise privacy concerns. To address these issues, we propose a novel approach for automatic synthetic data generation and introduce the \textbf{I}mplicit \textbf{P}ersonalized \textbf{Dialog}ue (\ipdialog) benchmark along with a training dataset, covering 10 tasks and 12 user attribute types. Additionally, we develop a systematic evaluation framework with four metrics to assess both attribute awareness and reasoning capabilities. We further propose five causal graphs to elucidate model reasoning pathways during implicit personalization. Extensive experiments yield insightful observations and prove the reliability of our dataset. Our dataset and code are available at \url{https://github.com/OpenCausaLab/IP-Dialog}.
\end{abstract}
\renewcommand*{\thefootnote}{\fnsymbol{footnote}}
\footnotetext[1]{Equal contribution.}
\footnotetext[2]{Corresponding author.}
\section{Introduction}
\begin{figure}[t]
    \centering
    \includegraphics[width=1\linewidth]{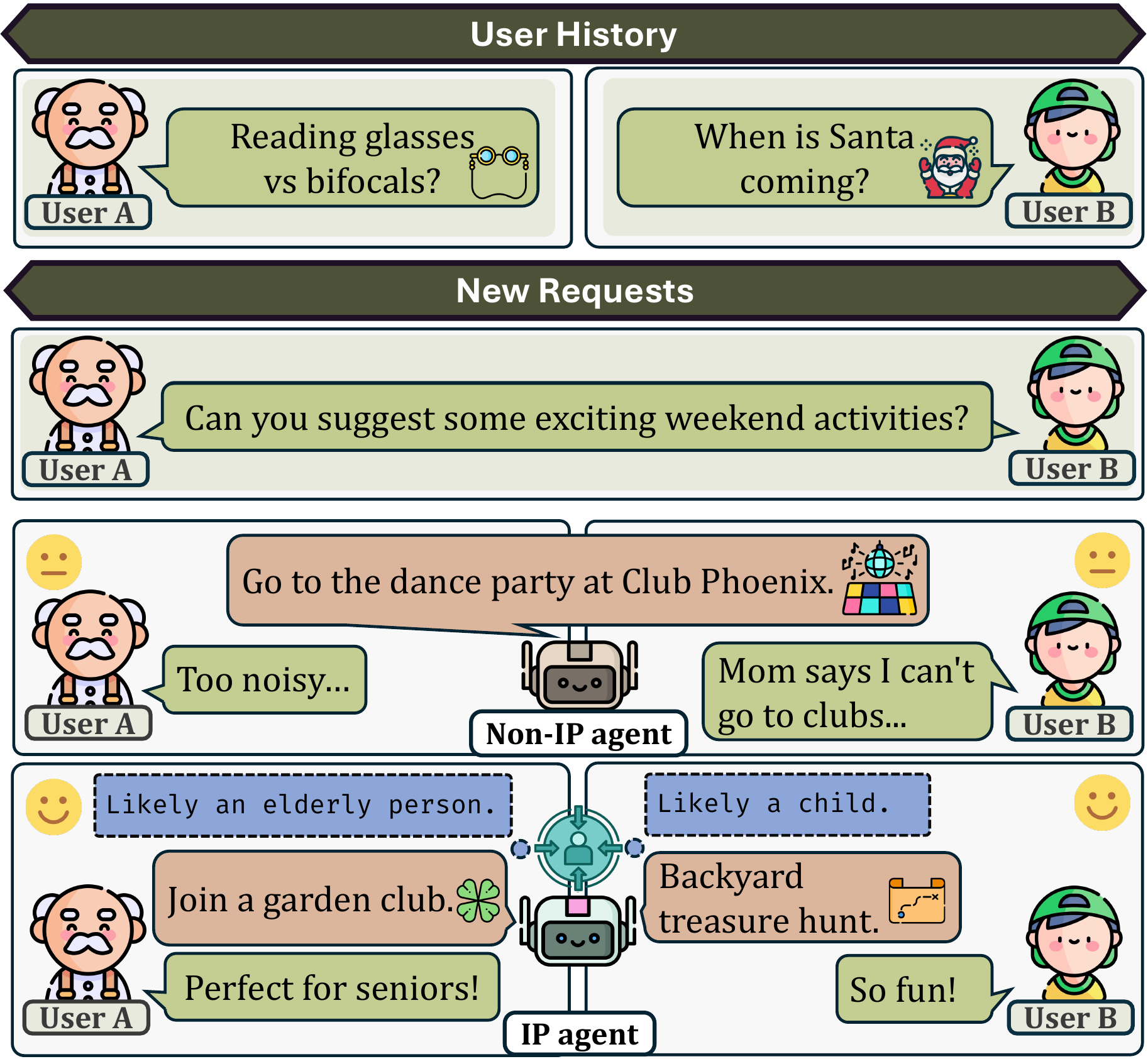}
    \caption{A comparative example of an AI agent with implicit personalization capability (IP agent) and one without (non-IP agent). The IP agent infers implicit user identities from dialogue history and generates customized responses accordingly.}
    \vspace{-4mm}
    \label{fig:example}
\end{figure}

Implicit personalization (IP)~\citep{flek-2020-returning, raharjana2021user, DBLP:journals/corr/abs-2405-14808}
, which involves tailoring responses based on inferred user characteristics without explicit user profiles, is crucial for enhancing the user experience in various AI-driven systems, including conversational agents~\citep{DBLP:conf/naacl/AnanthaVTLPC21,DBLP:journals/corr/abs-2305-09617,DBLP:conf/nips/ZhuangYWSZ23}, recommendation systems~\citep{DBLP:conf/emnlp/WangCLLL23}, and personalized content delivery~\citep{qian-etal-2024-tell}. In human-AI dialogues, user identities are implicitly embedded in the context of their inputs. These latent identities are vital in determining user preferences and shaping the expected AI responses~\citep{flek-2020-returning,raharjana2021user}. 
Figure \ref{fig:example} shows an example where AI agents answer questions based on user histories. An AI agent without IP capability may provide unsuitable suggestions, as it fails to infer user identities from the history. In contrast, an IP-capable agent can deliver personalized answers by recognizing users' latent identities (e.g., an elderly person or a child).
Such implicit personalization enables AI systems to provide more appropriate and engaging responses through a user-friendly approach.

However, no evaluation benchmarks or standards are available for IP, as publishing detailed user information causes privacy violation risks~\citep{DBLP:conf/uss/CarliniTWJHLRBS21,DBLP:conf/iclr/CarliniIJLTZ23}. Moreover, conventional manually labeled dataset construction approaches are prohibitively expensive and time-consuming. Considering the success of synthetic data~\citep{DBLP:conf/iclr/XuSZG0FTLJ24,DBLP:conf/iclr/Lou0XSAX0024,DBLP:conf/nips/YukhymenkoSVV24,10.5555/3666122.3668142}, we decide to utilize this advancement
and propose an automated data generation pipeline powered by state-of-the-art LLMs.
With this pipeline, we establish the \textbf{I}mplicit \textbf{P}ersonalized \textbf{Dialog}ue (\ipdialog) benchmark.

Our benchmark covers three carefully designed scenarios, encompassing 10 tasks with four distinct answer formats. We characterize users through 12 key attribute types 
(e.g., age, profession).
Each benchmark item consists of a user history for attribute inference and a user question that requires the model to incorporate the inferred attributes into its response. The user questions are generated through a multi-stage process: starting with 10 to 15 manually curated domains
(e.g., sports, education)
per task, generating 10 model-produced subjects per domain, and finally creating 10 user questions per subject that span diverse user attribute combinations. The user history is constructed iteratively, with each dialogue turn refined to reflect a single user attribute. The resulting dataset is divided into a training set (10,790 samples) and the \ipdialog~benchmark (1,000 samples). 

To systematically evaluate the IP capabilities of models, we establish a comprehensive evaluation framework comprising four primary metrics: two measuring attribute awareness 
and two evaluating attribute-based reasoning abilities. Furthermore, we propose five causal graphs to model how LLMs reason within the \ipdialog~task. These graphs range from a basic approach that disregards user attributes to more sophisticated reasoning pathways involving hidden attribute prediction and relevant attribute identification. Finally, we conduct extensive experiments across six models, yielding the following key findings:
\begin{enumerate}
  [itemsep=2pt,topsep=2pt,parsep=0pt]
    \item Models that excel at identifying relevant attribute types also demonstrate high accuracy in predicting the correct attribute values.
    \item \claude~achieves the best performance across all metrics. 
    Both \claude~and \gptf~have outperformed humans in solving IP tasks.
    \item 
    Tasks in the \textbf{behavior analysis} scenario, such as action prediction and preference inference, present the greatest challenge due to their dependence on complex psychological factors.
    \item The most effective reasoning pathway is \cotTsAs~
    which begins with inferring \textit{related attribute types}, followed by guessing \textit{related attributes} and finally providing the \textit{response}. \cotAs~serves as a viable alternative by directly inferring \textit{related attributes} before responding. Their high performance is mainly due to the precise and efficient attribute-related consideration process.
    Models with stronger IP capabilities show more resilience to variations in reasoning pathways.
    \item Supervised fine-tuning (SFT) significantly enhances the IP capability of \llamas~beyond all other models. 
    Models after SFT
    adapt well to unseen tasks with familiar answer formats but struggle with new formats. Moreover, SFT on a single reasoning pathway improves performance across other pathways.
\end{enumerate}

Our contributions are summarized as follows:
\begin{itemize}
  [itemsep=2pt,topsep=2pt,parsep=0pt]
    \item We design an efficient and highly controllable synthetic data methodology,
    providing solutions to data scarcity, privacy risks and evaluation challenges across various AI applications. 
    \item We introduce the \ipdialog~benchmark and the corresponding evaluation framework. To our knowledge, we are the first to evaluate the IP capabilities of LLMs in dialogue systems.
    \item We explore the impact of reasoning pathways on model performance in IP through five hypothesized causal graphs.
    \item Extensive experiments yield insightful observations and five key findings.
\end{itemize}

\section{Design of \ipdialog} 
Current AI-human dialogues can be conceptualized as consisting of a user historical dialogue (user history, $H$) and the current user request (user question, $Q$). The user history encapsulates the user's \textit{hidden attributes} $A$, which are not explicitly stated but can be inferred from past interactions. Implicit personalization (IP) in dialogues can be defined as a two-step process: first, inferring the \textit{related attributes} $A_s$ relevant to $Q$ from $H$, and then leveraging $A_s$ to generate personalized responses.
Following this definition, we construct the \ipdialog~benchmark, where each benchmark item consists of a task name, user history $H$, user question $Q$, \textit{related attributes} $A_s$, and the ground truth \textit{response}, which includes both \textit{analysis} and \textit{answer} components. 

\begin{table*}[t]
\centering
\footnotesize
\begin{tabular}{m{2.6cm}|m{12.5cm}}
\toprule
\textbf{Task name} & \multicolumn{1}{c}{\textbf{Definition}}\\
\specialrule{0.08em}{0pt}{0pt}
\rowcolor{morandiPink} \multicolumn{2}{c}{\rule{0pt}{2.ex}Recommendation System~\citep{DBLP:conf/ijcai/GuoTYLH17,DBLP:conf/kdd/LinQGDT0023}\rule[-1.ex]{0pt}{0pt}} \\
\specialrule{0.08em}{0pt}{2pt}
Recommendation (Rec, Open-ended) & \textbf{Definition}: Generate personalized item recommendations based on user attributes.~\citep{DBLP:conf/ijcai/GuoTYLH17}\newline
\textbf{Example}: I'm looking for high-performance sports gear that can support me during my workouts.\\
\specialrule{0.08em}{2pt}{2pt}
Ranking \newline(Rank, Ranking)&  \textbf{Definition}: Prioritize items based on user preferences and requirements.~\citep{DBLP:journals/corr/abs-1205-2618} \newline \textbf{Example}: Please rank the following online courses in order of suitability for me: A. online music course B. online painting course C. online cooking  course D. online language course\\
\specialrule{0.08em}{2pt}{2pt}
Filtering (Fil, Multiple choice) &  \textbf{Definition}: Select items that match specific user needs.~\citep{DBLP:journals/corr/abs-1708-05031} \newline
\textbf{Example}: Which of the following car models would appeal to my lifestyle: A. compact electric car B. rugged off-road SUV C. luxury convertible sports car D. practical minivan E. classic muscle car\\
\specialrule{0.08em}{2pt}{0pt}
\rowcolor{morandiGreen} \multicolumn{2}{c}{\rule{0pt}{2.ex}Behavior Analysis~\citep{morita1994information,6588663}\rule[-1.ex]{0pt}{0pt}} \\
\specialrule{0.08em}{0pt}{2pt}
Prediction (Pred, Binary choice) & \textbf{Definition}: Predict whether a user will take a specific action.~\citep{pi2019practice,
% ajzen2008attitudes,
megargee1976prediction} \newline 
\textbf{Example}: Predict whether I will attend a live basketball game this weekend.\\
\specialrule{0.08em}{2pt}{2pt}
Preference Inference (PI, Multiple choice)& \textbf{Definition}: Infer user's preference patterns by analyzing attribute-preference correlations.~\citep{yang2014modeling,kelly2003implicit} \newline
\textbf{Example}: What kind of exercise environment would I enjoy more? A. Group classes B. Home workouts C. Outdoor activities D. Gym workouts\\
\specialrule{0.08em}{2pt}{2pt}
Risk Detection (RD, Open-ended) & \textbf{Definition}: Identify potential risks based on user characteristics.~\citep{blanchard2011risk}
% ,wang2018neural,DBLP:conf/clef/BucurCD21} 
\newline
\textbf{Example}: What risks do you think I might face?\\
\specialrule{0.08em}{2pt}{2pt}
Intention Inference (II, Open-ended) & \textbf{Definition}: Uncover underlying motivations from ambiguous user requests.~\citep{kuo-chen-2023-zero} \newline
\textbf{Example}: Can you determine why I might want to explore specific athlete training routines?\\
\specialrule{0.08em}{2pt}{0pt}
\rowcolor{morandiBlue} \multicolumn{2}{c}{\rule{0pt}{2.ex}Action Guide~\citep{ajzen1985intentions}\rule[-1.ex]{0pt}{0pt}} \\
\specialrule{0.08em}{0pt}{2pt}
Advice (Adv, Open-ended) & \textbf{Definition}: Provide practical solutions to user problems. ~\citep{wang-torres-2022-helpful,kim2024advisorqa}\newline
\textbf{Example}: I'm interested in finding a way to incorporate sports into my routine. Any suggestions?\\
\specialrule{0.08em}{2pt}{2pt}
Decision (Dec, Binary choice) & \textbf{Definition}: Make yes/no suggestions for user decisions.~\citep{yaniv2004receiving,green2019principles} \newline
\textbf{Example}: Should I take up yoga classes to help manage stress and improve flexibility?\\
\specialrule{0.08em}{2pt}{2pt}
Convincing (Conv, Open-ended) & \textbf{Definition}: Develop persuasive arguments tailored to user characteristics.~\citep{zeng2024counterfactual,DBLP:journals/ker/Prakken06,DBLP:conf/lrec/YoshinoIMSS018} \newline 
\textbf{Example}: Convince me to try rock climbing.\\
\bottomrule
\end{tabular}
\caption{Task name (along with abbreviation and answer format), definitions, and examples of task questions.
}
\vspace{-4mm}
\label{tab:task_example}
\end{table*}

\subsection{User Attributes}
To comprehensively model user diversity, we design 12 attribute types that significantly influence users' needs, preferences, and behavior patterns. These attribute types are: age, gender, income level, profession, residence, Big Five personality traits, health status, and personal interests. 
Appendix \ref{app:design_detail_user} provides their corresponding attribute values.

\subsection{Tasks}
We categorize our 10 proposed tasks into three practical application scenarios: \textbf{Recommendation System}, \textbf{Behavior Analysis}, and \textbf{Action Guide}. To accommodate diverse task requirements, we define four distinct answer formats: \textbf{open-ended}, \textbf{ranking}, \textbf{multiple-choice}, and \textbf{binary-choice}. 
See Table \ref{tab:task_example} for task definitions.

\begin{figure*}[t]
   \centering
    \includegraphics[width=1\linewidth]{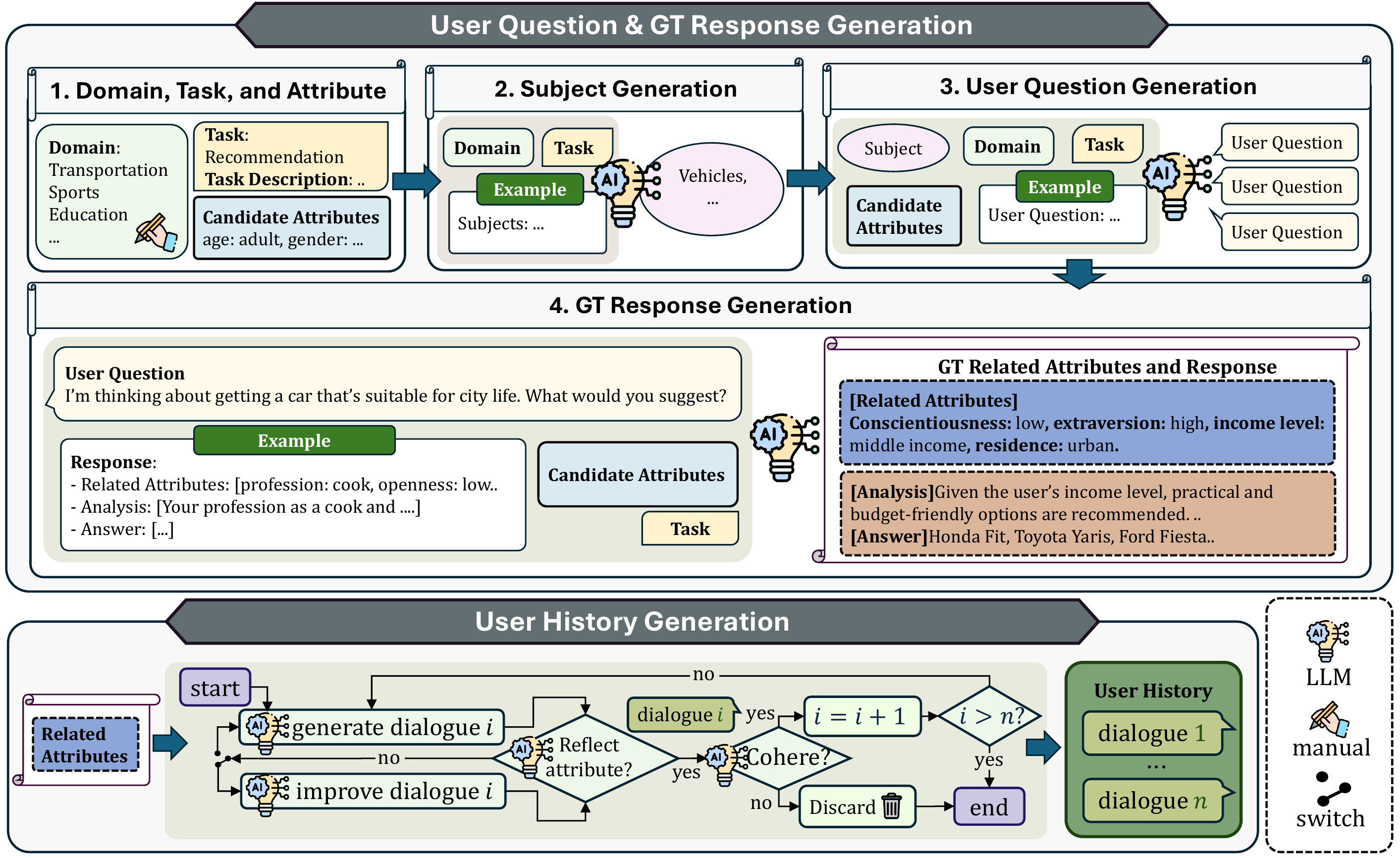}
    \caption{Construction pipeline of the \ipdialog~dataset. User questions and ground-truth (GT) responses are generated through: (1) Design domains, tasks, and attributes manually; (2) LLM generate subjects based on domain, task and manually designed examples with LLM; (3) Generate user questions based on subject, domain, task, manually designed examples and the candidate attributes from Section \ref{sec:usr_attribute_construct} with LLM; (4) Generate GT responses based on task, user question and candidate attributes with LLM. User history is generated based on the related attributes derived from (4). In each step, we generate a single attribute $i$. We introduce interactive checks and regeneration to ensure the attribute is reflected in the dialogue and the dialogue is coherent with all related attributes.}
    \label{fig:ques_construct}
    \vspace{-4mm}
\end{figure*}

%%%%%%%%%%%%%%%%%%%%%%%%%%%%%%%%%%%%%%%%%%%%%%%%%%%%%%%%%%%%%%%%%%%%%%%%%%%%%%

\section{Construction of \ipdialog}
\label{sec:const}

Figure~\ref{fig:ques_construct} illustrates our dataset generation pipeline.
We construct a total of 11,790 items, from which we randomly sample 1,000 items to form the \ipdialog~benchmark, ensuring efficiency and cost-effectiveness in evaluation. The remaining items constitute the training set. Detailed statistics and pseudo-code are given in Appendix \ref{app:detial_for_benchmark}. 

\subsection{User Attributes Construction}
\label{sec:usr_attribute_construct}
For each attribute type, we randomly select an attribute value to form a user's \textit{candidate attributes}. 
We implement specific constraints to prevent unrealistic attribute combinations, such as assigning ``\textit{retired}'' as a profession for a child.

\subsection{User Question Construction}
\label{sec:user_que}

For each task, we begin by manually selecting 10-15 most common domains. Next, we prompt \gptf~\citep{DBLP:journals/corr/abs-2303-08774} to generate 10 relevant subjects for each domain based on the task description. For each subject, we provide \gptf~with the \textit{candidate attributes} and instruct it to generate user questions.
To ensure high-quality generation, we include manually crafted examples as guidance. Once a subject or user question is generated, we sample 15 items for quality check. If any fails the check, we refine the prompts and regenerate. After generating user questions, \gptf~identifies the \textit{related attributes} from the \textit{candidate attributes}, performs \textit{analysis} on how these attributes influence the user's need, and finally generates the \textit{answer}. 

\subsection{User History Construction}
\label{sec:history_const}
We utilize \textit{related attribute}s to generate \textit{user history}. During construction, we find that generating history that reflects all \textit{related attributes} in one turn directly is challenging. Therefore, we design to generate history with $|\textit{related attributes}|$ steps and a check-and-refine procedure. At each step, \gptf~generates a single-round dialogue $i$ that implicitly reflects one related attribute $i$ from the \textit{related attributes}. Then, \gptf~verifies whether the generated dialogue can reflect the intended attribute. If not, the dialogue undergoes either improvement or regeneration: improvement refines the previously generated dialogue, while regeneration produces a new dialogue without referencing the previous one. These two strategies are alternated manually. They can help to balance the effectiveness of incremental refinement and the need to solve the situation when the prior generation is difficult to enhance. The check-and-refine cycle continues until the dialogue successfully reflects the intended attribute.
Once a dialogue $i$ reflecting attribute $i$ is successfully generated, we perform a coherence check to detect any conflicts\footnote{For example, dialogue $i$ ``\textbf{My grandkids} buy me a beautiful dress'' conflicts with the \textit{related attributes} \{gender:~female,~age: \textbf{child}\}, as a child cannot have grandchildren.} between \textit{related attributes} and dialogue $i$.
If the check fails, the intended user attributes will be removed from the dataset.
After passing this check, the process moves to the next step.

\section{Evaluation of IP Capability}
\label{sec:how_eval}

\subsection{Evaluation Framework for IP Ability}
Our evaluation framework systematically assesses IP in agent dialogue across three key dimensions:

\paragraph{Attribute Type Determination.}
Given a user history with information on
\textit{hidden attribute types} $T$ (e.g., $[$age,~health,~hobby$]$), 
an IP-capable model should identify which attribute types are most helpful to the current user question. To quantify this capability, we denote the predicted \textit{related attribute types} as $T_s$ and introduce \textit{attribute type F1} (\textit{ATF}):
{\fontsize{10pt}{12pt}
\begin{equation*}
\textit{ATF} = F_1(T_s) = \frac{2 \cdot \text{Precision}(T_s) \cdot \text{Recall}(T_s)}{\text{Precision}(T_s) + \text{Recall}(T_s)}.
\end{equation*}
}

\paragraph{Attribute Value Inference.}
After identifying the \textit{related attribute types}, models need to predict their corresponding attribute values correctly. 
Let $T_s^*$ denote the ground-truth \textit{related attribute types} (e.g., $[$age,~hobby$]$), and $A_s^*$ denote the ground-truth \textit{related attributes} (e.g., \{age:~child,~hobby:~music\}).
Each attribute type in $T_s^*$ corresponds to exactly one attribute value in $A_s^*$, so $|T_s^*| = |A_s^*|$.
We propose the \textit{relative value accuracy} (\textit{RVA}) score:
{\fontsize{10pt}{12pt}
\begin{equation*}
\begin{aligned}
\textit{RVA} &= \frac{|A_s \cap A_s^*|}{|T_s \cap T_s^*|} = 
% \frac{|A_s \cap A_s^*|}{|A_s^*|} \bigg/ \frac{|T_s \cap T_s^*|}{|T_s^*|}
\frac{|A_s \cap A_s^*|/|A_s^*|}{|T_s \cap T_s^*|/|T_s^*|}
= \frac{\text{Recall}(A_s)}{\text{Recall}(T_s)}.
\end{aligned}
\end{equation*} 
}
Among the correctly identified \textit{related attribute types}, the \textit{RVA} measures the proportion of their corresponding attribute values that are accurately predicted. 

\paragraph{Response Generation.} 
\label{para: accuracy_criteria} 
We evaluate response generation using both conventional metrics and LLM-based assessment. Conventional metrics, such as F1-score, offer efficient and deterministic evaluation, while LLM-based assessment enables customized evaluation as well as providing unified scores across different task formats.
For conventional metrics, we define \textit{task accuracy} as classification accuracy for \textbf{binary-choice} tasks, F1 score for \textbf{multiple-choice} tasks, Kendall's Tau coefficient for \textbf{ranking} tasks, and METEOR score for \textbf{open-ended} tasks.
For LLM-based assessment, we introduce \textit{\gptf-Score}, which uses \gptf~for evaluation. Referencing previous works on LLM-as-a-Judge~\citep{10.5555/3666122.3668142,DBLP:conf/icml/CuiY0YH0NXXL0024}, we define four key criteria: \textbf{conciseness} (0-1 points), \textbf{personalization} (0-4 points), \textbf{analysis quality} (0-4 points), and \textbf{answer accuracy} (0-5 points).
For the evaluation prompt, see Appendix \ref{app:gpt4o_score}.

\subsection{Reasoning Pathways for IP}
\label{sec:reason_pathway}
\begin{figure}[t]
    \centering
    \includegraphics[width=0.9\linewidth]{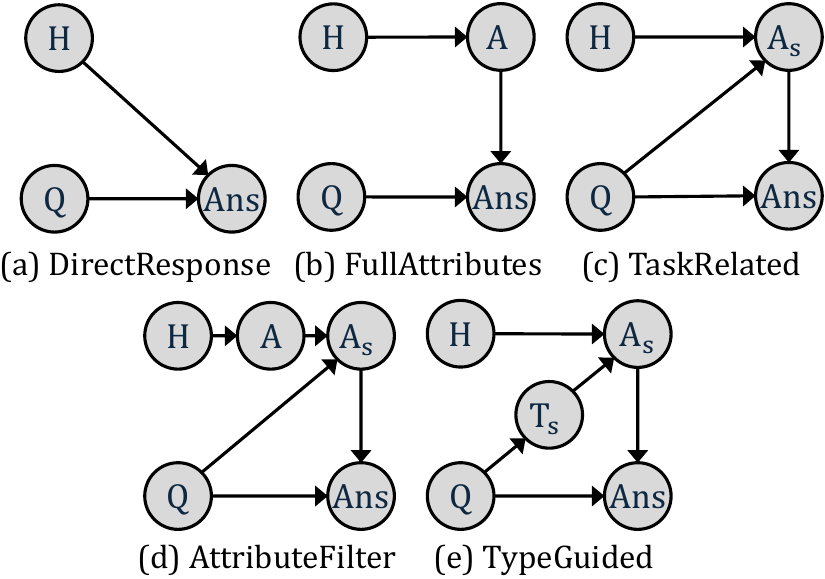}
    \caption{Five reasoning pathways represented as causal graphs. $H$ represents \textit{user history}, $Q$ denotes \textit{user question}, $A$ indicates \textit{hidden attributes}, $A_s$ refers to \textit{related attributes}, and $T_s$ is the \textit{related attribute types}.}
    \vspace{-4mm}
    \label{fig:reason_SCM}
\end{figure}

Next, we investigate the reasoning process of models on IP tasks. We formalize the five most common reasoning pathways
as causal graphs~\citep{pearl2009causality} and design their corresponding Chain-of-Thought (CoT) \citep{wei2022cot} prompts. Each pathway embodies a different hypothesis on how models should process user attributes. Shown in Figure \ref{fig:reason_SCM}, the five reasoning pathways are: (1) \basic~– the simplest approach, where the model generates a \textit{response} without explicitly considering the user attributes. (2) \cotA~– the model first predicts all \textit{hidden attributes} $A$ of the user, then leverage these attributes to generate the \textit{response}. (3) \cotAs~– the model directly identifies \textit{related attributes} $A_s$ before generating the \textit{response}. (4) \cotAAs~– the model first predicts \textit{hidden attributes} $A$, then extracts \textit{related attributes} $A_s$, and finally generates the \textit{response}. (5) \cotTsAs~– the model first infers \textit{related attribute types} $T_s$, then predicts specific \textit{related attributes} $A_s$,  and ultimately provides the \textit{response}.

\begin{table*}[t]
\footnotesize
\renewcommand
\arraystretch{1.0}
\centering
\setlength{\tabcolsep}{3pt}
\begin{tabular}{l|ccc|cccc|ccc|c}
\toprule
\multirow{4}{*}{\begin{tabular}{@{}c@{}}
\begin{tikzpicture}[overlay]
    % 放置文字在指定坐标
    \node[rotate=0] at (2.5,0.62) {Scenario};
    \draw (1.1, 0.335) -- (3.1,  0.335);
    \node[rotate=0] at (2.6,0.08) {Task};
    \draw (1.65,-0.205) -- (3.1,-0.205);
    \node[rotate=0] at (2.6,-0.35) {Answer}; % 旋转的文字
    \node[rotate=0] at (2.6,-0.6) {format}; % 旋转的文字
    \node at (0.5,0.08) {Method};
    % 画连接线
    \draw (0.5,0.9) -- (2.2,-0.75);
\end{tikzpicture}
\end{tabular}}
& \multicolumn{3}{c|}{Recommendation system} & \multicolumn{4}{c|}{Behavior analysis} & \multicolumn{3}{c|}{Action guide} & \multirow{4}{*}{Average}\\
\cmidrule{2-11}
      & ~~~Rec~~~ & Rank & Fil    &Pred   & PI      & RD   & II   & Adv  & Dec  & Conv  & \\
\cmidrule{2-11}
       & O   & R    & M     & B     & M       & O    & O    & O    & B    & O     &\\
\midrule
Random     & 0.00  & 4.29  & 11.00 & 50.00 & 19.25 & 0.00 & 0.00 & 0.00 & 50.00 & 0.00 & 13.45\\
\midrule
\gptomini  & 25.13 & 53.53 & 66.27 & 65.20 & 65.69 & 27.92 & 29.85 & 33.94 & 73.40 & \worstbox{23.63} & 46.46 \\
\gptf      &29.43 & \bestbox{65.07} & 64.06 & \bestusftbox{67.40} & 64.31 & 29.88 & 31.22 & \bestbox{37.72} & \bestusftbox{76.80} & 29.89 & 49.58\\
\claude    &\bestusftbox{31.55} & 61.98 & \bestusftbox{67.43} & 62.40 & \bestusftbox{67.71} & \bestusftbox{31.13} & \bestusftbox{31.81} & 36.75 & 75.00 & \bestusftbox{33.07} & \bestusftbox{49.88}\\
\llamal    &21.82 & \worstbox{42.33} & \worstbox{42.80} & 54.40 & 42.31 & \worstbox{21.98} & \worstbox{24.23} & \worstbox{26.00} & 59.60 & 25.81 & \worstbox{36.13} \\
\llamas    &\worstbox{21.46} & 42.68 & 49.25 & 56.80 & \worstbox{38.04} & 24.85 & 28.39 & 33.69 & 60.80 & 25.21 & 38.12\\
\qwen      &24.97& 52.57& 48.61& 63.80& 50.57& 23.11& 25.32& 32.18& 70.80& 25.58& 41.75\\
\midrule
Baseline Avg.       &25.73& 53.03& 56.40& 61.67& 54.77& 26.48& 28.47& 33.38& 69.40& 27.20& 43.65\\
\midrule
SFT-Full   &35.15 & 57.63 & \bestbox{69.68} & \bestbox{75.80} & \bestbox{71.02} & 47.15 & 38.96 & 36.26 & \bestbox{83.80} & \bestbox{36.29} & \bestbox{55.17}\\
SFT-w/o Rec-Fil-Dec & 32.70 & 58.53 & 61.07 & 70.60 & 69.36 & 47.59 & 39.32 & 35.34 & 80.20 & 35.70 & 53.04\\
SFT-w/o B  & \bestbox{35.56} & 59.14 & 67.87 & \worstbox{37.40} & 70.16 & \bestbox{48.28} & \bestbox{39.47} & 35.29 & \worstbox{8.60} & 35.14 & 43.69\\
\bottomrule
\end{tabular}
\caption{Average \textit{task accuracy} across all reasoning pathways. ``O'' represents \textbf{open-ended}, ``R'' represents \textbf{ranking}, ``M'' represents \textbf{multiple-choice}, and ``B'' represents \textbf{binary-choice}.
``Baseline Avg.'' stands for the average \textit{task accuracy} of the six non-fine-tuned baselines. ``SFT-Full'', ``SFT-w/o Rec-Fil-Dec'', and ``SFT-w/o B'' correspond to \llamas~fine-tuned on \textit{\textbf{Full}}, \textit{\textbf{w/o Rec-Fil-Dec}}, and \textit{\textbf{w/o B}} training datasets, respectively. For each task, we highlight the \bestbox{highest score}, the \worstbox{lowest score}, and the \bestusftbox{highest score among non-fine-tuned models}. Note that if the highest overall score is achieved by a non-fine-tuned model, only the blue highlight is used. }
\label{table:task_acc}
\vspace{-4mm}
\end{table*}

\section{Experiments}
We begin our experiments with model performance evaluation across three dimensions. Then, we investigate the influence of different reasoning pathways. After that, we analyze the effectiveness of supervised fine-tuning with our training set. Finally, we conduct automatic and human quality evaluation to prove the reliability of our synthetic dataset and its alignment to real-world user conversations.
\subsection{Setup}
\paragraph{Model.}
We evaluate six leading LLMs: GPT-4o~\citep{DBLP:journals/corr/abs-2303-08774}, \gptomini~\cite{openai2024o1},  \claude~\citep{claude2024}, \llamas, \llamal~\citep{meta_llama_2024}, and \qwen~\citep{qwen2.5}. 

\paragraph{Metric.}
We use the four metrics in Section \ref{sec:how_eval} for evaluation: \textit{attribute type F1} (\textit{ATF}), \textit{relative value accuracy} (\textit{RVA}), \textit{task accuracy}, and \textit{\gptf-Score}.
% to evaluate model performance on \ipdialog~benchmark.

\subsection{Performance Evaluation}
\begin{figure}[t]
    \centering
    \includegraphics[width=1\linewidth]{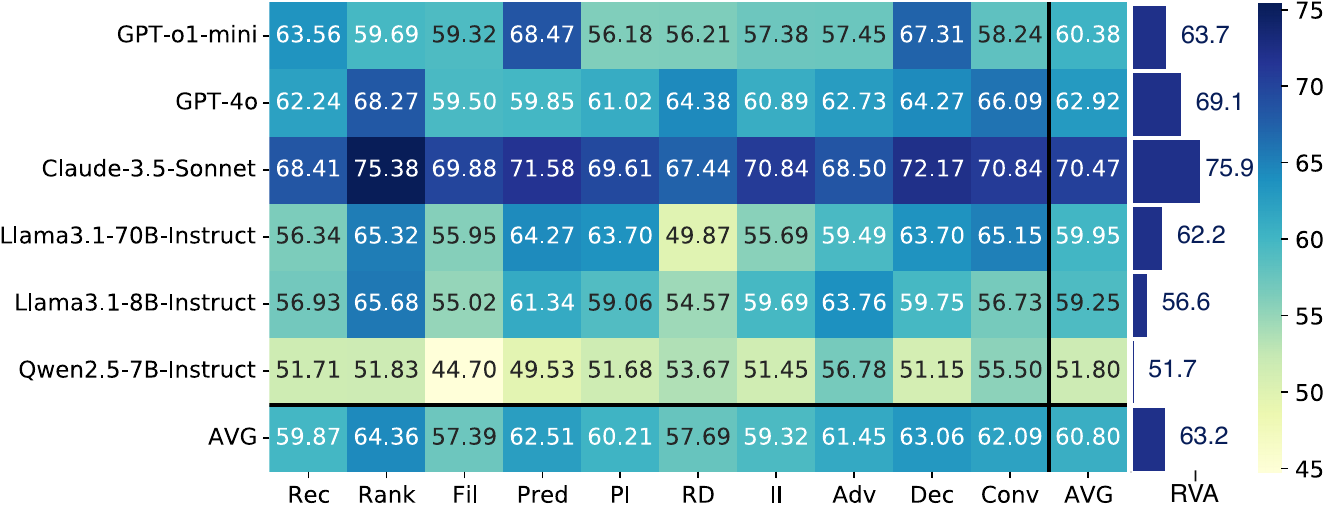}
    \vspace{-7mm}
    \caption{\textit{Attribute type F1} and \textit{relative value accuracy}. The heatmap illustrates the \textit{ATF} across models and tasks. The bar chart on the right shows the average \textit{RVA} for each model. The two metrics exhibit a strong positive correlation with Pearson's correlation of 0.957.}
    \vspace{-2mm}
    \label{fig:attr_performance}
\end{figure}

\paragraph{Attribute Performance.}
To assess model capabilities in determining and inferring attributes, we evaluate their average performance across three reasoning pathways: \cotAs, \cotAAs, and \cotTsAs. These pathways are selected because they all consider extracting the \textit{related attributes} from history explicitly ($H\rightarrow A_s$). Figure \ref{fig:attr_performance} presents the \textit{ATF} and \textit{RVA} results, it reveals that:
(1) \textbf{Strong positive correlation exists between \textit{ATF} and \textit{RVA}:} Models with higher \textit{ATF} also achieve higher \textit{RVA}, with Pearson's correlation coefficient reaching $0.957$. This suggests that strengthening either capability naturally possibly enhances the other.
(2) \textbf{Ranking is the easiest task. Filtering is the most challenging.}
(3) \textbf{\claude~is the top on both metrics.
}

\begin{figure}[t]
    \centering
    \includegraphics[width=0.95\linewidth]{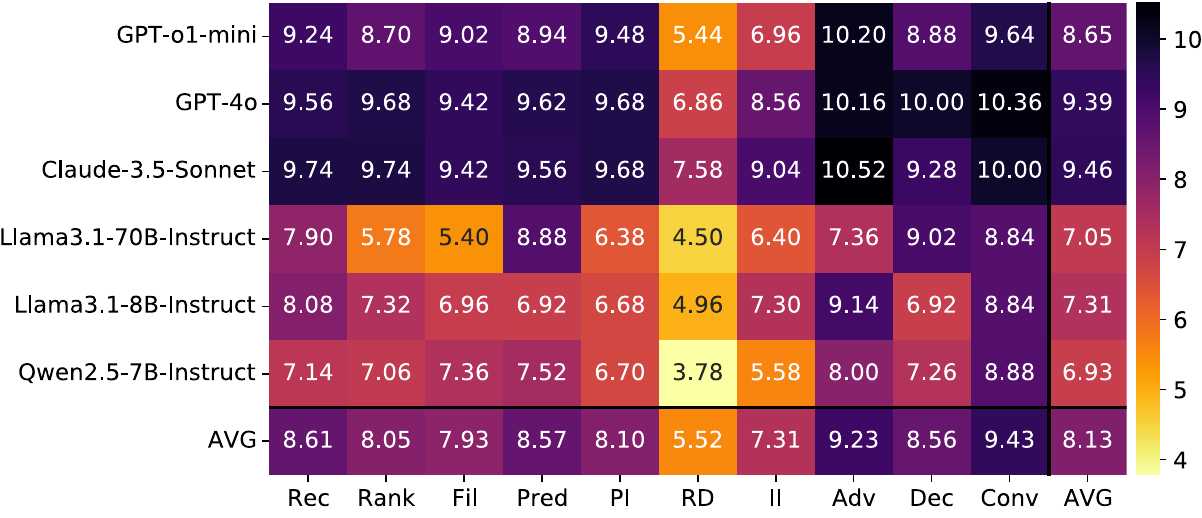}
    \vspace{-2mm}
    \caption{\textit{\gptf-Score} across models and tasks, averaged on all reasoning pathways. \gptf~scores model \textit{responses} from 0-14 based on criteria in Section \ref{para: accuracy_criteria}.}
    \vspace{-6mm}
    \label{fig:gpt4o_model_tasks}
\end{figure}

\paragraph{Task Accuracy.}
\label{sec:task_difficulty}
Table \ref{table:task_acc} reports the \textit{task accuracy} across models and tasks. We find that: (1) \textbf{All models perform significantly above random guessing,} indicating their fundamental IP capability. 
(2) \textbf{\claude~achieves the highest average \textit{task accuracy},} outperforming other models across most tasks.
(3) \textbf{A correlation emerges between attribute cognition and task performance:} Among the top three models, their ranking in \textit{task accuracy} (\claude~$>$ \gptf~$>$ \gptomini) 
aligns with their ranking in \textit{ATF}, suggesting that stronger attribute recognition contributes to task accuracy in high-performing models.

\begin{figure*}[t]
\centering
\subfigure[Reasoning pathway effectiveness. ]{
\label{fig:Cot_reason}
\begin{minipage}{0.66\textwidth}
\centering
\includegraphics[width=1\linewidth]{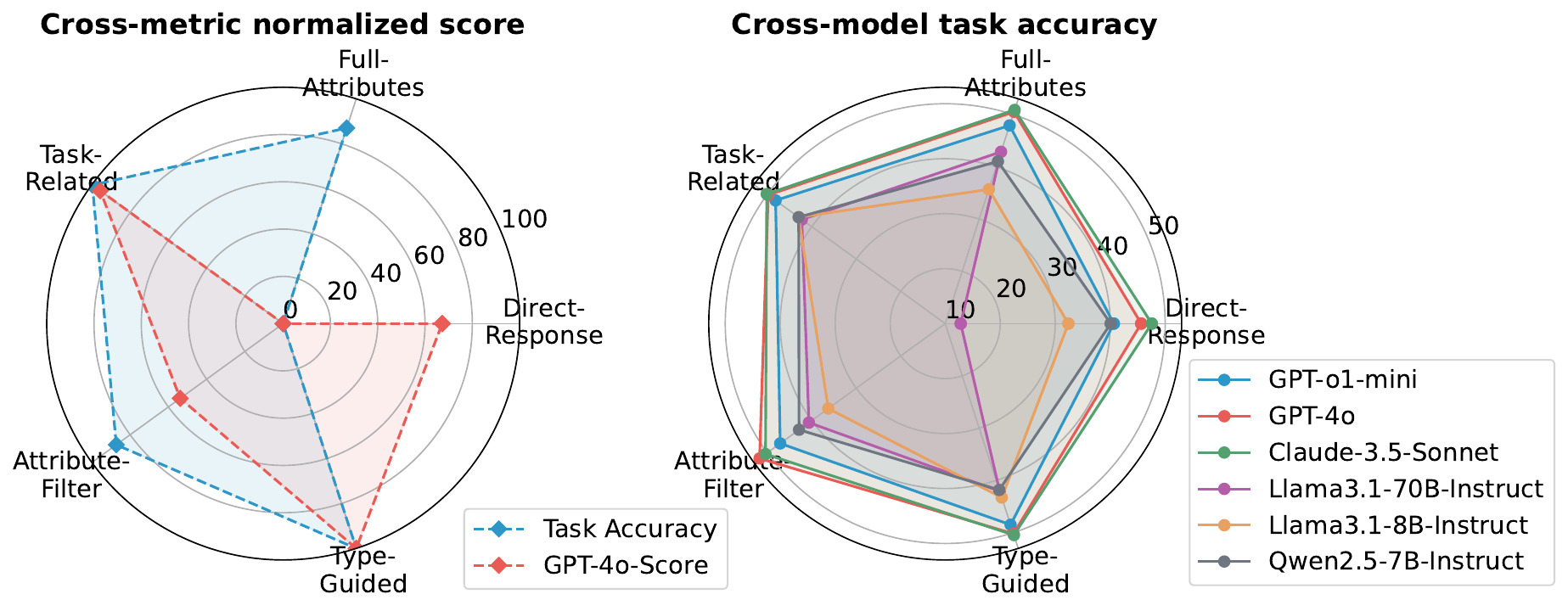}
% \vspace{0.5pt}
\end{minipage}
}
\hspace{2mm}
\subfigure[Average \textit{task accuracy} of \llamas~before and after SFT. ]{
\label{fig:pathway_task_accuracy_SFT}
\begin{minipage}{0.29\textwidth}
\centering
\includegraphics[width=1\linewidth]{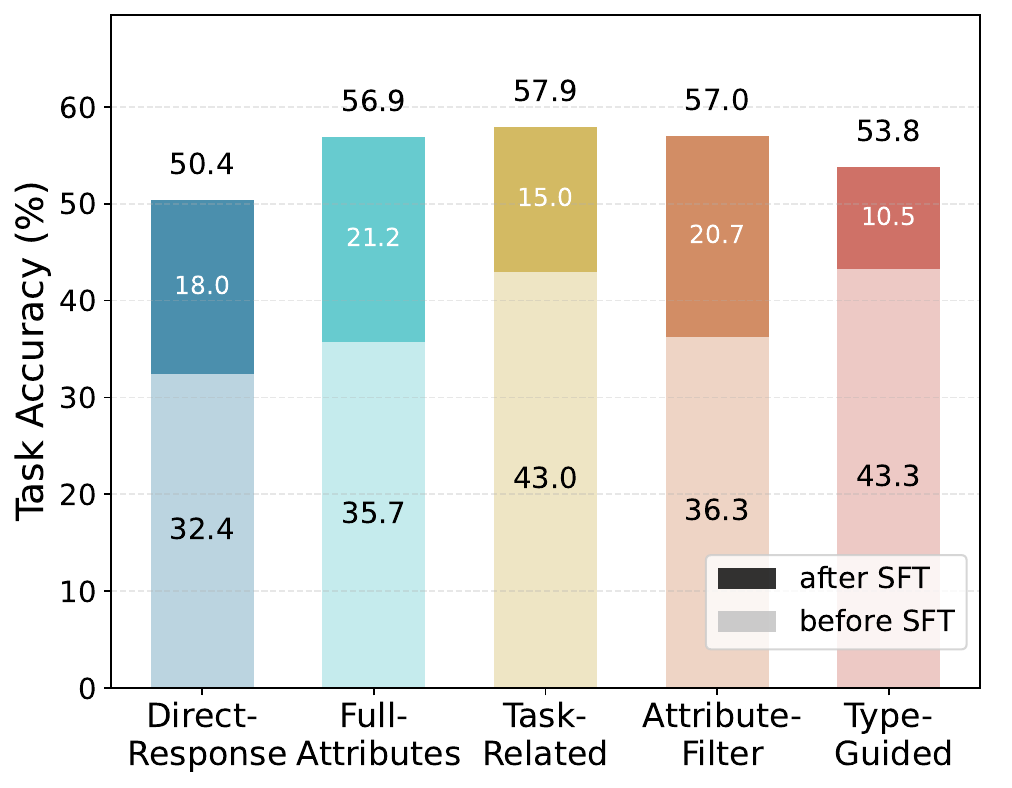}
\end{minipage}
}
\vspace{-4mm}
\caption{Reasoning pathway evaluation. (a) The left sub-figure compares the normalized average \textit{task accuracy} and \gptf-Score across five reasoning pathways, averaged on all models. The right sub-figure presents the average \textit{task accuracy} of models under each pathway. \cotTsAs~and \cotAs~demonstrate the highest effectiveness. (b) We use \cotAs~as the ground-truth reasoning pathway for SFT, considering its efficacy and conciseness.
}
\vspace{-4mm}
\label{fig:Cot_compare}
\end{figure*}

\paragraph{GPT-4o-Score.}

Shown in Figure \ref{fig:gpt4o_model_tasks},
from a model perspective, 
(1) \textbf{\gptf~and \claude~achieve the highest average \textit{\gptf-Score}s,} 
while (2) \textbf{the two Llama models often produce invalid or meaningless responses}, particularly in \llamal.
From a task perspective,
(1) \textbf{models generally perform well on \textbf{convincing} but struggle with \textbf{risk detection}.}
(2) \textbf{Among all scenarios, the difficulty ranking is: \textbf{behavior analysis}$>$\textbf{recommendation system}$>$\textbf{action guide}.}
This aligns with scenario characteristics:
\textbf{behavior analysis} requires understanding complex psychological factors; \textbf{recommendation system} focuses on more concrete matching; the subjective nature of \textbf{action guide} leads to conservative scoring by AI judges. Successfully solving these hard scenarios will have substantial model performance gains. Moreover, error analyses on hard scenarios can develop a deeper understanding of how models interpret human behavioral patterns.

\subsection{Influence of Reasoning Pathways}
\label{sec:comp_reason_path}

The left-hand side of Figure \ref{fig:Cot_reason} visualizes the average performance of the five reasoning pathways in \textit{task accuracy} and \textit{\gptf-Score}, following min-max normalization for each metric. 
Combined with the model-specific performance on the right-hand side, we find that:
(1) \textbf{\cotTsAs~consistently demonstrates superior performance across both metrics, followed closely by \cotAs.} This indicates that the extraction of \textit{related attributes} is crucial for effective IP reasoning.
(2) \textbf{\cotA~and \basic~exhibit high variance between \textit{task accuracy} and \textit{\gptf-Score}}, suggesting that certain pathways may perform inconsistently across different evaluation criteria.
(3) \textbf{The effectiveness of certain reasoning pathways appears highly dependent on the model's fundamental capabilities.} This is particularly evident in \basic, where weak models struggle significantly.
(4) \textbf{High-performing models demonstrate less dependency on specific reasoning pathways,} indicating greater robustness in handling implicit personalization tasks. However, their performance with \textbf{DirectResponse} remains significantly weaker than other pathways.

\subsection{SFT on the Training Set}

We fine-tune \llamas~using our training set of 10,790 items,
% (Section \ref{sec:const})
adopting the \cotAs~reasoning pathway. To further analyze the model adaptability across tasks and answer formats after SFT, we construct three datasets: (1) \textit{\textbf{Full}}: The original training dataset. (2) \textit{\textbf{w/o Rec-Fil-Dec}}: To evaluate the adaptability in unseen tasks, we exclude three tasks (\textbf{recommendation}, \textbf{filtering}, and \textbf{decision}) from \textit{\textbf{Full}}. (3) \textit{\textbf{w/o B}}: To assess the adaptability in unseen answer formats, we remove all \textbf{binary choice} tasks: \textbf{predicting} and \textbf{decision}.

We evaluate the fine-tuned models across all reasoning pathways and report the average \textit{task accuracy} in Table \ref{table:task_acc}: 
(1) \textbf{SFT-Full significantly outperforms the original model,} achieving the highest scores among all models, proving the effectiveness of SFT on improving IP capability.
(2) \textbf{SFT-w/o Rec-Fil-Dec generalizes well to unseen tasks,} indicating that the fine-tuned model can adapt well to new tasks with familiar answer formats.
(3) \textbf{SFT-w/o B exhibits severe performance degradation on binary-choice tasks,} failing in unfamiliar answer formats.
Analysis of its \textit{response}s shows that it tends to default to familiar formats from training rather than adopting the required new ones. To address this sensitivity, future training should incorporate more diverse formats.
(4) \textbf{Training with \cotAs~pathway enhances performance across other reasoning pathways,} demonstrating adaptability in reasoning patterns (Figure \ref{fig:pathway_task_accuracy_SFT}). 

\subsection{Automatic Quality Evaluation} 
\label{sec:auto_qua}
\begin{figure}[t]
% \centering
\subfigure[Semantic and lexical diversity.]{
\label{fig:diversity}
\begin{minipage}{0.28\textwidth}
\centering
\includegraphics[width=1.\linewidth]{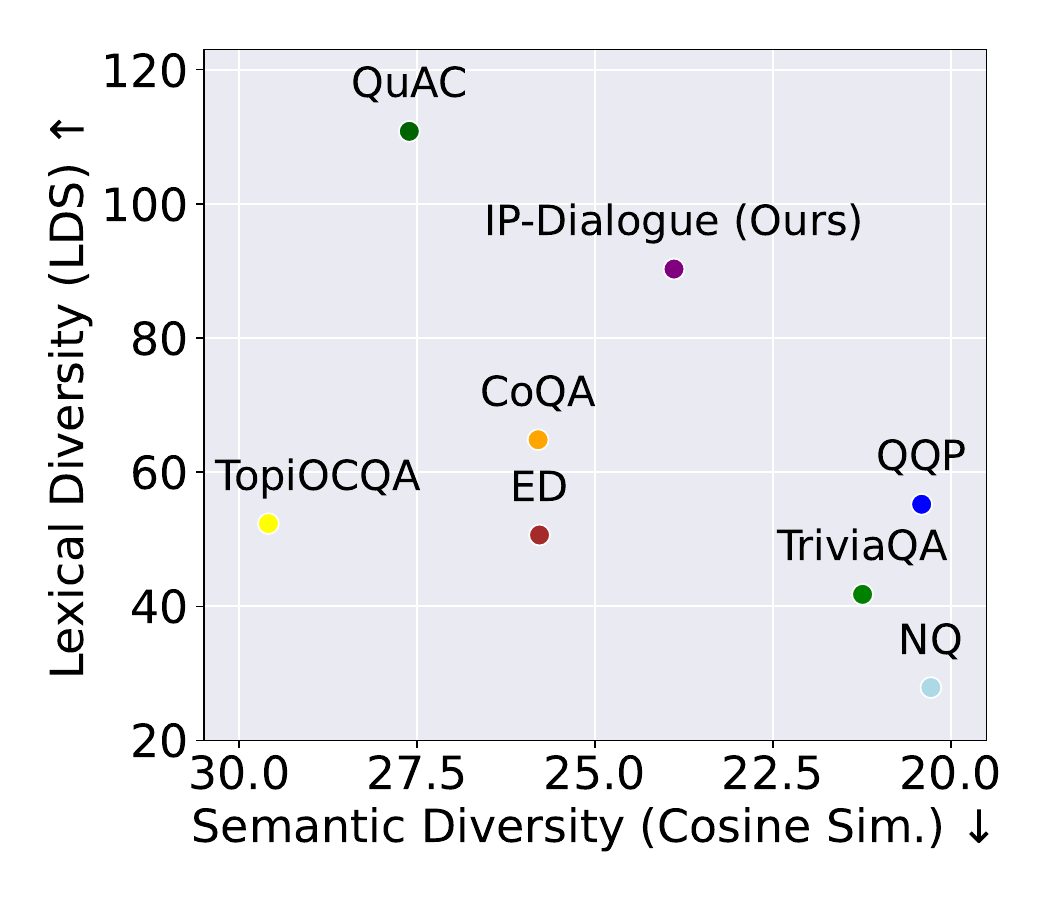}
\end{minipage}
}
% \hspace{1.mm}
\subfigure[Perplexity.]{
\begin{minipage}{0.13\textwidth}
\footnotesize
\centering
\setlength{\tabcolsep}{3pt}
% \vspace{0.2cm} 
\begin{tabular}{ccccccccc}
\toprule
Dataset & PPL$\downarrow$\\
\midrule
ED& 618.80\\
QQP& 113.26\\
NQ& 85.40\\
TriviaQA&65.56\\
QuAC & 422.43\\
CoQA & 1773.04\\
TopiOCQA& 599.72\\
\ipdialog &\textbf{45.14}\\
\bottomrule
\end{tabular}
% \resizebox{0.5\textwidth}{!}{
\vspace{0.15cm} 
% }
\label{tab:data_perplexity}
% \end{table*}
\end{minipage}
}
\\
\subfigure[Consistency.]{
\label{fig:consistency}
\begin{minipage}{0.49\textwidth}
\centering
\vspace{-0.35cm} 
\includegraphics[width=1\linewidth]{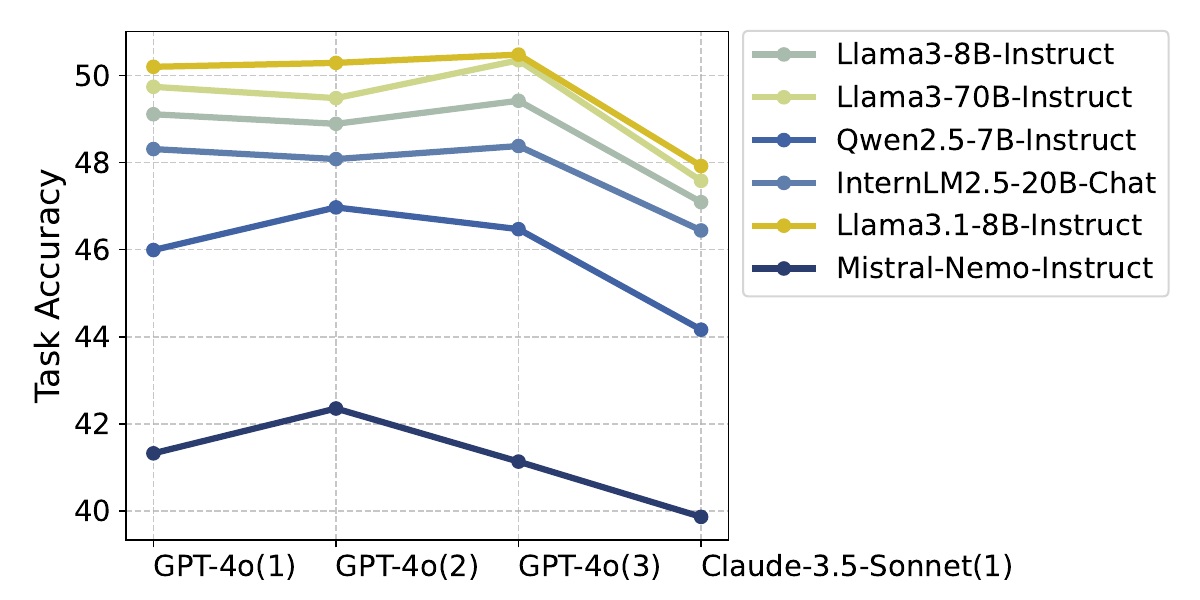}
\end{minipage}
}
\vspace{-3mm}
\caption[Automatic quality analysis]{Automatic quality analysis. \ipdialog~shows high semantic and lexical diversity (a) and superior linguistic fluency (b). Additionally, our dataset generation method achieves stable performance assessments across various generation models (c).}
\vspace{-2mm}
\label{fig:quality_auto}
\end{figure}
We conduct three automatic analyses. For more details, see Appendix \ref{app:auto_qual_eval}.

(1) \textbf{Diversity}: We use NV-Embed-v2 \citep{lee2024nv} for embeddings of each question, and then calculate the average cosine similarity. A lower cosine similarity indicates greater semantic diversity. We employ MATTR \citep{covington2010mattr}, MTLD \citep{mccarthy2005mtld}, and HD-D \citep{mccarthy2010hd-d}. We normalize and average these three metrics for a final lexical diversity score. Shown in Figure \ref{fig:diversity}, \ipdialog~achieves leading semantic and lexical diversity.

(2) \textbf{Fluency}: We use \llamas~for the perplexity score. With the lowest PPL score, \ipdialog~exhibits high fluency (Table \ref{tab:data_perplexity}). 

(3) \textbf{Consistency}: To assess benchmark reliability, we examine whether performance rankings remain stable across different dataset generation models. Figure \ref{fig:consistency} confirms this consistency, as the ranking of \textit{task accuracy} remains robust, validating the reliability of our benchmark.

%%%%%%%%%%%%%%%%%%%%%%%%%%%%%%%%%%%%%%%%%%%%%%%%%%%%%%%%%%%%%%%%%%%%%%%%%%%
\subsection{Human Study}
\label{sec:human_per}
We conduct human studies to evaluate human performance and dataset quality (Appendix \ref{app:human_study}). The setting of annotator number is shown in Table \ref{tab:human_setting}.

\begin{table}[t]
\centering
\footnotesize
% \resizebox{0.4\textwidth}{!}{
\begin{tabular}{m{3.25cm}m{0.8cm}m{0.2cm}m{1.1cm}}
\toprule
Experiment & \multicolumn{1}{c}{Samples} & \multicolumn{1}{c}{Ann.} & Ann. per sample\\
\midrule
Attribute Inference      &  \multicolumn{1}{c}{100} &  \multicolumn{1}{c}{2}  &  \multicolumn{1}{c}{1}  \\
Task Accuracy      & \multicolumn{1}{c}{50}  &  \multicolumn{1}{c}{4}  & \multicolumn{1}{c}{2} \\
\midrule
Fidelity & \multicolumn{1}{c}{200} & \multicolumn{1}{c}{6} & \multicolumn{1}{c}{3} \\
Attribute-Dialogue Align. & \multicolumn{1}{c}{200} & \multicolumn{1}{c}{4} & \multicolumn{1}{c}{2} \\
Attribute-Response Align. & \multicolumn{1}{c}{200} & \multicolumn{1}{c}{4} & \multicolumn{1}{c}{2}\\
\bottomrule
\end{tabular}
% }
\caption{Human study setting. ``Ann.'' means annotators.}
\vspace{-4mm}
\label{tab:human_setting}
\end{table}

For human performance, in
% \paragraph{Human Performance}
(1) \textbf{Attribute Inference Accuracy}, annotators are tasked with inferring attribute types and values from each of the historical dialogues under a predefined set of possible attributes in Table \ref{tab:attribute_table}. 
In (2) \textbf{Task Accuracy}, annotators answer questions based on a set of \textit{candidate attributes}. For both experiments, \textbf{human annotators perform better than or comparably with Llama3-70B-Instruct, but worse than \gptf~and \claude.}
This reflects our tasks' high cognitive demands: IP-tasks require advanced reading comprehension, attention to subtle details, and extensive world knowledge, where LLMs have more advantages than humans. The breakthrough enables the development of reliable LLM-driven personalization services to reduce human efforts.

For quality analysis, 
(1) \textbf{Fidelity}: A Turing test yields an accuracy of 52.2\%, indicating that our dataset is nearly indistinguishable from human-generated data. (2) \textbf{Attribute-dialogue Alignment}: 
Human reviewers find that 92.0\% of utterances accurately reflect their corresponding ground-truth attributes, demonstrating the high reliability of our dataset. 
(3) \textbf{Attribute-response Alignment}: 
Annotators assess the consistency between \textit{responses} and \textit{related attributes}, as well as the logical coherence of \textit{analysis}. Among the evaluated samples, 91.9\% meet these assessment standards, confirming the dataset's robustness. 

\section{Related Work}
\paragraph{Personalization on Implicit Inference.}
Recently, \citet{DBLP:journals/corr/abs-2405-14808} introduced the concept of implicit personalization (IP), which involves inferring user backgrounds from their queries and tailoring responses accordingly. Current research related to IP is limited. 
A possible related research direction is user intention understanding \citep{DBLP:conf/sigir/QuYCTZQ18,DBLP:conf/um/CaiC20,DBLP:conf/acl/KuoC23,qian-etal-2024-tell}, but they rely on the explicit user answers. Table \ref{tab:comparison}, Appendix \ref{app:relat_com} presents a detailed comparison between existing work and ours.

\paragraph{Personalization on Explicit Information and Historical Sequences.}
While distinct from implicit personalization, other research in personalization offers valuable insights. One line of research focuses on explicit information-based personalization~\citet{hovy-2015-demographic,DBLP:conf/aaai/JangLHOSLSKL22,he2024cosenhancingpersonalizationmitigating}.
Another line of research focuses on analyzing the historical sequences of users to predict future behaviors~\citep{sasaki-etal-2018-predicting}. Among them, LaMP \citep{DBLP:conf/acl/SalemiMBZ24} aggregates 7 tasks for LLM personalization, serving as dataset in many follow-up studies \citep{DBLP:journals/corr/abs-2406-02888,liu2024llmspersonaplug,DBLP:journals/corr/abs-2407-11016,DBLP:journals/corr/abs-2402-04401}

\section{Conclusion and Discussions}
We provide a comprehensive view of implicit personalization. Through an efficient and controllable generation pipeline, we create the \ipdialog~benchmark alongside a training dataset. 
We develop an evaluation framework featuring four primary metrics and design hypothesized causal graphs to investigate potential reasoning pathways in IP. With extensive experiments, we provide insightful findings and prove our dataset's reliability.

\section{Limitations}
Though we make our best effort to include as many tasks and user attributes as possible, some of the values are not covered. The limited user attribute design is due to trade-offs between synthetic cost and diversity coverage, as expanding attributes like neurodivergence or intersectional identities demand exponential efforts. 
Moreover, though our experiments have proved the reliability and fidelity of our datasets, we admit that there could be a potential discrepancy between synthetic dialogues and the real-world user conversations. Due to huge human efforts and time costs to gather such real-world data, we leave this problem to our future work.
Finally, we must acknowledge the potential risks associated with the advance of IP technology. IP systems might cause societal stereotypes or biases. To mitigate these risks, we suggest incorporating bias control techniques and restrictions to avoid stereotypes and discrimination.
%%%%%%%%%%%%%%%%%%%%%%%%%%%%%%%%%%%%%%%%%%%%%%%%%%%%%%%%%%%%%%
%%%%%%%%%%%%%%%%%%%%%%%%%%%%%%%%%%%%%%%%%%%%%%%%%%%%%%%%%
\bibliography{custom}

\clearpage
\listofappendices

\appendix

\vspace{6mm}
%%%%%%%%%%%%%%%%%%%%%%%%%%%%%%%User attributes and values.%%%%%%%%%%%%%%%%%%%%%%%%%%%%%%%
\begin{table}[htb]
\centering
\footnotesize
\begin{tabular}{l|l}
\toprule
\textbf{Attribute Type} & \textbf{Attribute Values} \\
\midrule
\multirow{3}{*}{age} & child, teenager, \\
                     &young adult, adult, \\
                     &middle-aged, elderly \\
\hline
income level & low, middle, high \\
\hline
\multirow{8}{*}{profession}& student, teacher, \\
                           & sales worker, office clerk, \\
                           & factory worker, \\
                           & software engineer,  \\
                           & healthcare worker, driver, \\
                           & farmer, cook, cleaner,\\
                           & journalist, retiree, \\
                           & entrepreneur \\
\hline
openness & high, low \\
\hline
conscientiousness & high, low \\
\hline
extraversion & high, low \\
\hline
agreeableness & high, low \\
\hline
neuroticism & high, low \\
\hline
residence & urban, rural \\
\hline
gender & male, female \\
\hline
 \multirow{2}{*}{health}& healthy, minor issue, \\
                        & disabled \\
\hline
\multirow{4}{*}{hobby}& reading, cooking, \\
                      & photography, sports, \\
                      & gaming, movies, travel, \\
                      & music, crafts \\
\bottomrule
\end{tabular}
\caption{User attributes and values. These attributes are carefully selected to characterize users and their diverse needs and preferences.}
\label{tab:attribute_table}
\end{table}
%%%%%%%%%%%%%%%%%%%%%%%%%%%%%%%User attributes and values.%%%%%%%%%%%%%%%%%%%%%%%%%%%%%%%
\section{Comparison with other works}
\addcontentsline{loa}{section}{\numberline{A}Comparison with other works}
\label{app:relat_com}
We present a detailed comparison between our work and the previous datasets in Table \ref{tab:comparison}.

%%%%%%%%%%%%%%%%%%%%%%%%%%%%%%%Comparison between IP-Dialog and existing datasets.%%%%%%%%%%%%%%%%%%%%%%%%%%%%%%%
\begin{table*}[h]
\centering
\setlength{\tabcolsep}{3pt}
\footnotesize
\begin{tabular}{l|c|c|c|c}
\toprule
\multirow{1}{*}{\textbf{Dataset}} & \multirow{1}{*}{\textbf{Source}} & \textbf{Context Type} & \textbf{User Attributes} & \textbf{Implicit Inference}\\
\midrule
LaMP {\small\citep{DBLP:conf/acl/SalemiMBZ24}} & Aggregation & Sequence & $\times$ & $\times$ \\
NQ {\small\citep{kwiatkowski2019natural}} & Crowdsourced & QA & $\times$ & $\times$\\
GlobalOpinionQA {\small\citep{durmus2023measuring}} & Survey & QA & \checkmark & $\times$ \\
GIE-Bench {\small\citep{DBLP:journals/corr/abs-2406-14903}} & Synthetic & QA & \checkmark & $\times$ \\
SynthPAI {\small\citep{yukhymenko2024synthetic}} & Synthetic & Comment & \checkmark & $\times$ \\
EmpatheticDialogues {\small\citep{DBLP:conf/acl/RashkinSLB19}} & Crowdsourced & Dialogue & $\times$ & $\times$ \\
IN3 {\small\citep{qian-etal-2024-tell}} & Synthetic & Dialogue & $\times$ & \checkmark \\
PersuasionForGood {\small\citep{wang2019persuasion}} & Crowdsourced & Dialogue & \checkmark & $\times$ \\
TopDial {\small\citep{wang2023target}} & Synthetic & Dialogue & \checkmark & $\times$ \\
\textbf{IP-Dialog (Ours)} & Synthetic & Dialogue & \checkmark & \checkmark \\
\bottomrule
\end{tabular}
\caption{Comparison between IP-Dialog and existing datasets. \textbf{Context Type}: primary format of the data (QA, dialogue, or preference sequence). \textbf{User Attributes}: whether the dataset includes user characteristics (e.g., income level, profession). \textbf{Implicit Inference}: whether the dataset requires reasoning from implicit information in context.}
\label{tab:comparison}
\end{table*}
%%%%%%%%%%%%%%%%%%%%%%%%%%%%%%%Comparison between IP-Dialog and existing datasets.%%%%%%%%%%%%%%%%%%%%%%%%%%%%%%%
%%%%%%%%%%%%%%%%%%%%%%%%%%%%%%%Statistics of \ipdialog. %%%%%%%%%%%%%%%%%%%%%%%%%%%%%%%
\begin{table*}[htb]
\centering
\footnotesize
% \begin{tabularx}{0.5\textwidth}{>{\hsize=0.5\hsize\centering\arraybackslash}X>{\hsize=0.3\hsize\centering\arraybackslash}X>{\hsize=0.2\hsize\centering\arraybackslash}X}
\begin{tabular}{lccccl}
\toprule
\textbf{Task} & \textbf{\#Domain} & \textbf{\#Subject} & \textbf{\#Sample} & \textbf{$|A_s|$} & \textbf{Len(H/Q)}\\
\specialrule{0.08em}{0pt}{0pt}
\rowcolor{morandiPink} \multicolumn{6}{c}{Recommendation System} \\
\specialrule{0.08em}{0pt}{2pt}
Recommendation& 14& 65 & 100 & 3.46& 177.11/17.43\\
Ranking & 14& 65 & 100 & 4.68&240.09/30.28\\
Filtering & 14& 62 & 100 & 3.21&164.53/41.3\\
\specialrule{0.08em}{2pt}{0pt}
\rowcolor{morandiGreen} \multicolumn{6}{c}{Behavior Analysis} \\
\specialrule{0.08em}{0pt}{2pt}
Prediction &15& 81 & 100 & 4.92&255.58/13.54\\
Preference Inference  & 17& 77 & 100 & 4.02&206.31/24.41\\
Risk Detection & -& -&100 & 5.0&256.71/8.0\\
Intention Inference& 12& 63 & 100 & 4.47&240.19/14.44\\
\specialrule{0.08em}{2pt}{0pt}
\rowcolor{morandiBlue} \multicolumn{6}{c}{Action Guide} \\
\specialrule{0.08em}{0pt}{2pt}
Advice& 12& 65 & 100 & 4.36&223.62/18.26\\
Decision & 12& 55 & 100 & 4.4&223.7/16.7\\
Convincing & 15& 69 & 100 & 4.42&243.52/14.3\\
\specialrule{0.08em}{2pt}{2pt}
\textbf{Average} & 14& 67 & 100 & 4.29& 223.14/19.87\\
\bottomrule
\end{tabular}
\caption{Statistics of \ipdialog. Each row shows the number of domains and subjects, the number of samples, the average number of related attributes ($|A_s|$), and the average length (in words) of history ($H$) and question ($Q$) for each task.}
\vspace{-4mm}
\label{tab:task_statistic_concise}
\end{table*}
%%%%%%%%%%%%%%%%%%%%%%%%%%%%%%%Statistics of \ipdialog. %%%%%%%%%%%%%%%%%%%%%%%%%%%%%%%

%%%%%%%%%%%%%%%%%%%%%%%%%%%%%%%%%%%%%%%%%%%%%%%%%%%%%%%%%%%%%%%%%%%%%%%%%%%%%%%%%%%%%%%%%%%%%%%%%%%%%%%%%%%%%%%%%%%
% Design Details
%%%%%%%%%%%%%%%%%%%%%%%%%%%%%%%%%%%%%%%%%%%%%%%%%%%%%%%%%%%%%%%%%%%%%%%%%%%%%%%%%%%%%%%%%%%%%%%%%%%%%%%%%%%%%%%%%%%
\section{Design Details}
\addcontentsline{loa}{section}{\numberline{B}Design Details}
\label{app:design_detail}

\subsection{User Attributes}
\label{app:design_detail_user}
\addcontentsline{loa}{subsection}{\numberline{B.1}User Attributes}
We provide the designed 12 attribute types and values in Table~\ref{tab:attribute_table}.

\subsection{Tasks}
\label{app:design_detail_task}
\addcontentsline{loa}{subsection}{\numberline{B.1}Tasks}
% Table \ref{tab:task_example} shows the answer format, definition, and example of tasks.
% examples with their corresponding answer formats. Table \ref{tab:task_def} summarizes the task definitions. 
We further detail our task design considerations and contributions below.

\paragraph{Recommendation System.} 
Recently, there is an increasing focus on leveraging large language models to improve recommendation systems \citep{DBLP:journals/www/WuZQWGSQZZLXC24,bao2023tallrec,harte2023leveraging}. 
Most current recommendation systems utilize user historical preference series for personalization~\citep{gao2023chat,christakopoulou2023large,DBLP:conf/acl/SalemiMBZ24}. While prior LLM research has explored conversational agents in recommendation systems~\citep{DBLP:conf/acl/LiuWNWCL20,dao2024broadening}, the potential relationships between user dialogues, implicit attributes, and latent interests remain largely unexplored. We propose that leveraging implicit user information presents a promising approach for enhancing recommendation quality. This approach could address several persistent challenges in recommendation systems, including the cold start problem~\citep{lika2014facing,lam2008addressing}, lack of recommendation diversity~\citep{kunaver2017diversity,zhang2008avoiding}, and the limitation in recognizing the potential needs of users~\citep{DBLP:conf/recsys/WangLLMWGZHBBCC24}. By analyzing implicit user attributes, LLMs can identify potential user needs and suggest relevant items without requiring explicit preferences. Subsequently, by leveraging these LLM-generated elements, the system can expand the recommendation results by discovering similar items in the database, delivering convenient, personalized, and rich recommendations to users.

\paragraph{Behavior Analysis.}
Behavior analysis serves as a fundamental cornerstone for improving user-centric services, such as content recommendations and preference-based customization. While traditional methods in behavior analysis typically rely on extensive user data, LLMs can leverage their intrinsic knowledge about the relationship between user attributes and behavior patterns to generate analytical insights. 
To comprehensively evaluate this capability, we design four representative tasks that cover different aspects of user behavior understanding. Among them, \textbf{intention inference}~\citep{kuo-chen-2023-zero,qian-etal-2024-tell} has been studied before. 
However, previous research relies on interactive dialogue, in which the agent asks the user for more specific detail~\citep{qian-etal-2024-tell,kuo-chen-2023-zero}. Developing a system that automatically infers user intent without explicit questioning would greatly enhance user convenience. 

\paragraph{Action Guide.}
\textbf{Action guide} aims to transform user intentions into concrete actions~\citep{ajzen1985intentions} through three complementary elements: generating practical solutions (\textbf{advice}), conducting decision analysis (\textbf{decision}), and facilitating behavior change (\textbf{convincing}). This scenario integrates informational, analytical, and motivational aspects of guidance to bridge the gap between knowledge acquisition and action implementation. Successfully bridging this gap is critical for personalized LLMs. 
%%%%%%%%%%%%%%%%%%%%%%%%%%%%%%%Examples of task questions with their corresponding answer formats.%%%%%%%%%%%%%%%%%%%%%%%%%%%%%%%
%%%%%%%%%%%%%%%%%%%%%%%%%%%%%%%%%%%%%%%%%%%%%%%%%%%%%%%%%%%%%%%%%%%%%%%%%%%%%%%%%%%%%%%%%%%%%%%%%%%%%%%%%%%%%%%%%%%
% Design Details
%%%%%%%%%%%%%%%%%%%%%%%%%%%%%%%%%%%%%%%%%%%%%%%%%%%%%%%%%%%%%%%%%%%%%%%%%%%%%%%%%%%%%%%%%%%%%%%%%%%%%%%%%%%%%%%%%%%

%%%%%%%%%%%%%%%%%%%%%%%%%%%%%%%%%%%%%%%%%%%%%%%%%%%%%%%%%%%%%%%%%%%%%%%%%%%%%%%%%%%%%%%%%%%%%%%%%%%%%%%%%%%%%%%%%%%
% Construction Details
%%%%%%%%%%%%%%%%%%%%%%%%%%%%%%%%%%%%%%%%%%%%%%%%%%%%%%%%%%%%%%%%%%%%%%%%%%%%%%%%%%%%%%%%%%%%%%%%%%%%%%%%%%%%%%%%%%%
\section{Construction Details}
\addcontentsline{loa}{section}{\numberline{C}Construction Details}
\label{app:detial_for_benchmark}
The statistics of \ipdialog~is shown in Table \ref{tab:task_statistic_concise}. The pseudo-code of user question generation and user history generation is shown in Algorithm \ref{alg:ip-dialog} and \ref{alg:ip-dialog2}. While our tasks were specifically designed to require user background information for appropriate responses, we acknowledge varying degrees of context-dependency across scenarios. In most cases, historical context significantly impacts the ground truth responses of our dataset. However, a very small subset of examples may exhibit low dependency on historical context, particularly in:
\begin{itemize}
    \item Binary or multiple-choice questions with limited answer options.
    \item Questions with strong inherent constraints that naturally narrow potential responses.
    \item Requests where objective reasoning dominates over personalization needs.
\end{itemize}
We deliberately included such instances to evaluate whether personalization systems can discern when contextual information is necessary versus when it isn't relevant. Real-world applications naturally contain questions with these varying personalization requirements, and our dataset reflects this authentic distribution. More generation details are explained below.

\begin{algorithm*}
\caption{User Question Construction}
\label{alg:ip-dialog}
\begin{algorithmic}[1]

\Require 
    \Statex \textit{domains} \Comment{Dict of domains for each task}
    \Statex \textit{tasks} \Comment{List of tasks with name, description and requirements}
    \Statex \textit{subject\_prompt} \Comment{Subject generation prompt}
    \Statex \textit{user\_question\_prompt} \Comment{User question generation prompt}
    \Statex \textit{GT\_prompt} \Comment{Attribute/response generation prompt}
\Ensure \textit{QA\_items}

\Statex
\State \# Generate candidate user attributes
\State $\textit{candidate\_attributes\_dataset} \gets \text{attribute\_generator}()$

\Statex
\State \# User question construction
\State $\textit{QA\_items} \gets [\ ]$
\For{each \textit{task} in \textit{tasks}} \Comment{length of tasks is 10}
    \State $\textit{candidate\_attributes\_loader} \gets \text{create\_iterator}(\textit{candidate\_attributes\_dataset})$
    \For{each \textit{domain} in \textit{domains}[\textit{task.name}]} \Comment{10-15 domains}
        \State $\textit{subjects} \gets \text{GPT4o}(\textit{subject\_prompt}(\textit{task}, \textit{domain}))$
        \For{each \textit{subject} in \textit{subjects}} \Comment{10 subjects}
            \For{$k \gets 1$ to $3$}
                \State $\textit{candidate\_attributes} \gets \textit{candidate\_attributes\_loader}.\text{next}()$
                \State $\textit{user\_questions} \gets \text{GPT4o}(\textit{user\_question\_prompt}(\textit{task}, \textit{domain}, \textit{subject},$\\\hspace{\algorithmicindent}$ \textit{candidate\_attributes}))$
                \For{each \textit{user\_question} in \textit{user\_questions}} \Comment{3 questions}
                    \State $\textit{related\_attributes}, \textit{analysis}, \textit{answer} \gets \text{GPT4o}(\textit{GT\_prompt}(\textit{task}, \textit{user\_question},$\\\hspace{\algorithmicindent}$ \textit{candidate\_attributes}))$
                    \State $\textit{QA\_items}.\text{append}((\textit{task}, \textit{domain}, \textit{subject}, \textit{user\_question}, \textit{related\_attributes}, $\\\hspace{\algorithmicindent}$\textit{analysis}, \textit{answer}))$
                \EndFor
            \EndFor
        \EndFor
    \EndFor
\EndFor

\State \Return $\textit{QA\_items}$

\end{algorithmic}
\end{algorithm*}
\begin{algorithm*}
\caption{User History Construction}
\label{alg:ip-dialog2}
\begin{algorithmic}[1]

\Require 
    \Statex \textit{regen\_improve\_list} \Comment{Regeneration/improvement strategies}
    \Statex \textit{dialog\_gen\_prompt} \Comment{Dialogue generation prompt}
    \Statex \textit{dialog\_improve\_prompt} \Comment{Dialogue improvement prompt}
    \Statex \textit{attri\_dialog\_align\_prompt} \Comment{Alignment examination prompt}
    \Statex \textit{dialog\_consistency\_prompt} \Comment{Consistency check prompt}
    \Statex \textit{related\_attributes\_dataset}  \Comment{Dataset of related attributes generated from Algorithm 1}
    \Statex \textit{QA\_items}  \Comment{Question-Answer items from Algorithm 1}
\Ensure \textit{IP\_dialog\_dataset}
\Statex

\State \# Extract unique related\_attributes combinations from QA items
\State $\textit{related\_attributes\_dataset} \gets \text{set}([\textit{QA\_item$[-2]$ for } \textit{QA\_item} \text{ in } \textit{QA\_items}])$
\Statex

\State \# User history construction
\State $\textit{item\_dialogues} \gets [\ ]$
\For{each \textit{related\_attributes} in \textit{related\_attributes\_dataset}}
    \State $\textit{dialogues} \gets [\ ],\ \textit{dialogue} \gets \text{""}$
    \For{$i, \textit{related\_attribute}$ in \textit{enumerate}(\textit{related\_attributes})} \Comment{Generate dialogue per attribute}
        \State $\textit{dialogue} \gets \text{GPT4o}(\textit{dialog\_gen\_prompt}(\textit{related\_attribute}, \textit{dialogue}))$
        \For{$j \gets 1$ to $31$} \Comment{Try up to 31 times}
            \State $\textit{reflected} \gets \text{GPT4o}(\textit{attri\_dialog\_align\_prompt}(\textit{dialogue}, \textit{related\_attribute}))$
            \If{$\textit{reflected}$ or $j = 31$}
                \State \textbf{break}
            \Else
                \If{$\textit{regen\_improve\_list}[j] = \text{"regeneration"}$}
                    \State $\textit{dialogue} \gets \text{GPT4o}(\textit{dialog\_gen\_prompt}(\textit{related\_attribute}, \textit{dialogue}))$
                \ElsIf{$\textit{regen\_improve\_list}[j] = \text{"improvement"}$}
                    \State $\textit{dialogue} \gets \text{GPT4o}(\textit{dialog\_improve\_prompt}(\textit{dialogue}, \textit{related\_attribute}))$
                \EndIf
            \EndIf
        \EndFor
        
        \If{not \textit{reflected}} 
            \State \text{discard this \textit{related\_attributes} combination}
        \EndIf
        \State $\textit{conflict} \gets \text{GPT4o}(\textit{dialog\_consistency\_prompt}(\textit{dialogue}, \textit{related\_attributes}))$
        \If{$\textit{conflict}$}
            \State \text{discard this \textit{related\_attributes} combination}
        \EndIf
        \State $\textit{dialogues}.\text{append}(\textit{dialogue})$
    \EndFor
    \State $\textit{item\_dialogues}.\text{append}((\textit{dialogues}, \textit{related\_attributes}))$
\EndFor

\Statex
\State \# Map dialogues to QA items
\State $\textit{IP\_dialog\_dataset} \gets \text{map\_dialogues\_to\_QA\_items}(\textit{item\_dialogues}, \textit{QA\_items})$
\State \Return $\textit{IP\_dialog\_dataset}$

\end{algorithmic}
\end{algorithm*}

\subsection{User Question Construction}
\addcontentsline{loa}{subsection}{\numberline{C.1}User Question Construction}
For ground truth (GT) answer generation and model evaluation, we limit \textit{related attributes} to no more than 5 to reduce complexity and improve accuracy assessment. During the construction of domains and subjects for user questions, the \textbf{risk detection} task stands as an exception, as it consists solely of user attributes without domain and subject distinctions. 

The generation process involves multiple specialized prompts (prompts for subject, user question, and GT related attribute and response) presented below.
% We present the generation details alongside illustrations of these specialized prompts. 
Within these prompt illustrations, the content enclosed in \{\} varies dynamically during generation based on specific tasks, domains, and contextual parameters. Sample values are shown in \{\} to aid comprehension.
% For subject and user question generation, we utilize the prompts detailed below. 
Note that during the generation of user questions, we utilized 3 user questions for each subject and user attribute candidate combination. 
\newpage
% The ground truth \textit{response} generation prompt is also illustrated below.
% \newpage
%%%%%%%%%%%%Prompt for Subject Generation%%%%%%%%%%%%%%%%%
% \begin{figure}[t!]
% \centering
\begin{tcolorbox}[
  enhanced,
  % breakable,
  title={\small Subject Generation},
  attach boxed title to top center={yshift=-3mm,yshifttext=-1mm},
  colback=gray!3, 
  colframe=black, 
  colbacktitle=gray!80!blue!70, 
  coltitle=white,
  fonttitle=\bfseries,
  boxed title style={size=small,colframe=black},
  separator sign={},
]
\small
\texttt{Generate 10 subjects on the domain: \{Sports\}. These subjects are for the task: \{decision\}. The task requires the agent to \{make a decision for a user on whether to perform an action or not.\}\\
The agent will leverage the user's attributes to answer the request related to the subject in a personalized way. The user's attributes include: age, income\_level, profession, openness, conscientiousness, extraversion, agreeableness, neuroticism, residence(urban/rural), gender, health(healthy/minor\_issue/disabled), hobby\\
---Important Note---\\
- Make your generated subject as diverse as possible. To cover as much as possible, with the greatest possible differences between categories.\\
- The subject should be applicable to all user attributes, and the subject should not include any hints about attributes.\\
- The length of the subject should be less than 5 words.\\
- Output your generated subjects in the format of:\\
Output subjects: \\
- subject 1\\
- subject 2\\
- ...\\
---Example Subject---\\
Suppose you are asked to generate subjects on the domain: \{Education\},\\
Output subjects: \\
\{- whether to pursue a graduate degree\\
- acceptance of university offer\\
- choice of major\}\\
---Your Generated Subject---\\
Now list your generated subject on the domain: \{Sports\}\\
Output subjects: }
\end{tcolorbox}
% \vspace{-4mm}
% \caption{Prompt for subject generation.}
% \vspace{-4mm}
% \label{fig:prompt_sub_gen}
% \end{figure}
%%%%%%%%%%%%Prompt for Subject Generation%%%%%%%%%%%%%%%%%

%%%%%%%%%%%%Prompt for User Question Generation%%%%%%%%%%%%%%%%%
% \begin{figure}[ht]
% \centering
\begin{tcolorbox}[
  enhanced,
  % breakable,
  title={\small User Question Generation},
  attach boxed title to top center={yshift=-3mm,yshifttext=-1mm},
  colback=gray!3,  
  colframe=black,  
  colbacktitle=gray!80!blue!70,  
  coltitle=white,
  fonttitle=\bfseries,
  boxed title style={size=small,colframe=black},
  separator sign={},
]
\small
\texttt{You are a request-generation engine. Your mission is to generate 3 requests on the subject: \{Sports\}-\{join a local sports league\}. These requests are for the task: \{decision\}. The task requires an agent to \{make a decision for a user on whether to perform an action or not.\}\\
The agent will leverage the user's attributes to answer the request in a personalized way. \\
---Important Note---\\
- You should generate requests in a first-person tone.\\
- Make your generated requests type as diverse as possible. Avoid generating similar requests.\\
- Do not mention words such as best fit my unique profile, or any other hints about attributes.\\
- \{Give your answer after the user request. Your user request should not contain any hints about the attributes. It should be hard to make a decision.\}\\
- Output your generated requests in the format of:\\
Output requests: \\
- request 1 [Answer: ...]\\
- request 2 [Answer: ...]\\
- ...\\
---Example Requests---\\
Suppose you are asked to generate request on subject: \{Education\}-\{whether to pursue a graduate degree\},\\
User Attributes: \{\{'age': 'adult', 'gender': 'female', 'income\_level': 'middle\_income', 'profession': 'cook', 'openness': 'low', 'conscientiousness': 'high', 'extraversion': 'high', 'agreeableness': 'low', 'neuroticism': 'high', 'residence': 'rural', 'health': 'minor\_issue', 'hobby': 'cooking'\}\}\\
Output requests: \\
\{Should I pursue a graduate degree in music? [Related Attributes: no music background, high conscientiousness, low openness, rural residence, minor health issue. Answer: No. Reason: Based on your current situation, it may not be the best choice for you.]\}\\
---Your Generated Requests---\\
Now list your generated requests on subject: \{Sports\}-\{join a local sports league\}\\
User Attributes:\\
\{\{'age': 'young\_adult', 'gender': 'female', 'income\_level': 'low\_income', 'profession': 'factory\_worker', 'openness': 'low', 'conscientiousness': 'low', 'extraversion': 'low', 'agreeableness': 'low', 'neuroticism': 'low', 'residence': 'rural', 'health': 'healthy', 'hobby': 'cooking'\}\}\\
Output requests: }
\end{tcolorbox}
% \vspace{-4mm}
% \caption{Prompt for user question generation.}
% \vspace{-4mm}
% \label{fig:prompt_request_gen}
% \end{figure}
%%%%%%%%%%%%Prompt for User Question Generation%%%%%%%%%%%%%%%%%

%%%%%%%%%%%%Prompt for GT Related Attribute and Response Generation%%%%%%%%%%%%%%%%%
% \begin{figure}[htbp]
% \centering
\begin{tcolorbox}[
  enhanced,
  % breakable,
  title={\parbox{0.8\linewidth}{\centering \small Related Attribute and Response Generation}},
  attach boxed title to top center={yshift=-3mm,yshifttext=-1mm},
  colback=gray!3,  
  colframe=black,  
  colbacktitle=gray!80!blue!70,  
  coltitle=white,
  fonttitle=\bfseries,
  boxed title style={size=small,colframe=black},
  separator sign={},
]
\small
\texttt{You are a helpful agent for the task: \{decision\}. The task requires agent to \{make a decision for a user on whether to perform an action or not.\}\\
You will leverage the user's attributes to respond to the request in a personalized way. The user request you need to respond to is: \{Should I join a local soccer league for fun and socializing?\}\\
---Important Note---\\
- First, list the user attributes you will use to answer the request in Attributes: [..]. Then, provide the reasoning process for your answer in Analysis: [..]. Finally, present your formatted answer in Answer: [..].\\
- You should select the user attributes that are most relevant to the request. Use no more than 5 attributes. The attributes should be selected from the given user attribute group.\\
- Your reasoning should be concise and clear.\\
- \{Your answer should be either yes or no.\}\\
- Output your response in the format below, do not omit the [] in your response:\\
Output: \\
- Attributes: [..].\\
- Analysis: [..].\\
- Answer: [..].\\
---Example---\\
User Attribute Group: \{\{'age': 'adult', 
'gender': 'female', 'income\_level': 'midd-\\
le\_income', 'profession': 'cook', 'openne-\\
ss': 'low', 'conscientiousness': 'high', 
'extraversion': 'high', 'agreeableness': 'low', 'neuroticism': 'high', 'residence': 'rural', 'health': 'minor\_issue', 'hobby': 'cooking'\}\}\\
\{User: Should I pursue a graduate degree in music?\\
- Attributes: [profession: cook, openness: low, conscientiousness: high, residence: rural].\\
- Analysis: [As a cook with low openness, pursuing a graduate degree in music represents a significant departure from the user's current profession and comfort zone. High conscientiousness suggests dedication to current work. Rural residence could limit access to music education opportunities.]\\
- Answer: [no]\}\\
---Your Response---\\
User Attribute Group: \{\{'age': 'young\_adu-\\
lt', 'gender': 'female', 'income\_level': 'low\_income', 'profession': 'factory\_work-\\er', 'openness': 'low', 'conscientiousnes-\\s': 'low', 'extraversion': 'low', 'agree-\\ableness': 'low', 'neuroticism': 'low', 'residence': 'rural', 'health': 'healthy', 'hobby': 'cooking'\}\}\\
User: \{Should I join a local soccer league for fun and socializing?\}\\
Output: }
\end{tcolorbox}
% \vspace{-4mm}
% \caption{Prompt for ground truth related attributes and response generation.}
% \vspace{-4mm}
% \label{fig:prompt_ans_gen}
% \end{figure}
%%%%%%%%%%%%Prompt for GT Related Attribute and Response Generation%%%%%%%%%%%%%%%%%

\subsection{User History Construction}
\addcontentsline{loa}{subsection}{\numberline{C.2}User History Construction}

To prepare attributes for user history generation, we extract and aggregate \textit{related attributes} mentioned in the ground truth \textit{responses} of user questions to form a collection of attribute combinations. Duplicate combinations are consolidated to ensure uniqueness within the set. Subsequently, for each unique \textit{related attributes} combination in this set, we generate corresponding user history dialogues. 

For the history generation, we implement an iterative approach consisting of 31 generation-examination iterations per step. An example of our manually designed improvement(i)/regeneration(r) choices 
is i-i-i-t-i-i-...-i-t-i-t.
Generated dialogue in each step that fail to meet our consistency criteria is discarded. As such cases only account for a small portion of our generation results, removal proves more efficient than remediation. 

The history dialogue generation process encompasses four prompt types: (1) initial history dialogue generation (and regeneration) for step 0 and step 1+, (2) attribute-alignment examination, (3) iterative improvement described, and (4) consistency verification. The generated history dialogues are paired with user questions sharing the same \textit{related attributes} to construct the final dataset.

After dataset construction, we compute cosine similarity scores between user questions across all samples, constructing a subset where all pairwise similarity scores fall below a threshold of 0.6. Then, we randomly sample 1,000 instances from this filtered subset to form the \ipdialog~benchmark, with the remaining samples comprising the training set.

%%%%%%%%%%%%Prompt for Dialogue Generation/Regeneration (Step 0)%%%%%%%%%%%%%%%%%
% \begin{figure}[ht]
% \centering
\begin{tcolorbox}[
  enhanced,
  % breakable,
  title={\parbox{0.8\linewidth}{\centering \small Dialogue Generation/Regeneration (Step 0)}},
  attach boxed title to top center={yshift=-3mm,yshifttext=-1mm},
  colback=white,
  colframe=black,
  colbacktitle=gray!80!red!70,
  coltitle=white,
  fonttitle=\bfseries,
  boxed title style={size=small,colframe=black},
  separator sign={},
  % width=0.8\textwidth,
]
\small
\texttt{Generate a single-round dialogue between a user and AI to implicitly reflect the following user attribute:\\
\{extraversion: low\}\\
\\
Guidelines:\\
1. Do not explicitly mention the user attributes.\\
2. The generated dialogue should implicitly reflect the user's attribute, allowing it to be inferred from the context.\\
3. Ensure the dialogue remains natural, as if between a user and an AI assistant.\\
\\
Format the dialogue strictly as follows:\\
User: [Brief user message that implicitly reflects the given attribute]\\
AI: [Brief AI response that naturally responds to the user's message without referencing the user's attributes]}
\end{tcolorbox}
% \vspace{-4mm}
% \caption{Prompt for dialogue generation/regeneration (Step 0).}
% \vspace{-4mm}
% \label{fig:prompt_his_gen}
% \end{figure}
%%%%%%%%%%%%Prompt for Dialogue Generation/Regeneration (Step 0)%%%%%%%%%%%%%%%%%

%%%%%%%%%%%%Prompt for Dialogue Generation/Regeneration (Step 1+)%%%%%%%%%%%%%%%%%
% \begin{figure}[htbp]
% \centering
\begin{tcolorbox}[
  enhanced,
  % breakable,
  title={\parbox{0.8\linewidth}{\centering \small Dialogue Generation/Regeneration (Step 1+)}},
  attach boxed title to top center={yshift=-3mm,yshifttext=-1mm},
  colback=white,
  colframe=black,
  colbacktitle=gray!80!red!70,
  coltitle=white,
  fonttitle=\bfseries,
  boxed title style={size=small,colframe=black},
  separator sign={},
  % width=0.8\textwidth,
]
\small
\texttt{Based on the following dialogue:\\
\{User: Can you tell me about some quiet places to visit for a relaxing afternoon?\\
AI: Sure! How about a scenic park or a cozy library nearby?\}\\
\\
Continue the dialogue for one more round. The continued single-round dialogue should be between a user and AI to implicitly reflect the following user attribute:\\
\{agreeableness: low\}\\
\\
Guidelines:\\
1. Do not explicitly mention the user attributes.\\
2. The generated dialogue should implicitly reflect the user's attribute, allowing it to be inferred from the context.\\
3. Ensure the dialogue remains natural, as if between a user and an AI assistant.\\
\\
Format the dialogue strictly as follows:\\
User: [Brief user message that implicitly reflects the given attribute]\\
AI: [Brief AI response that naturally responds to the user's message without referencing the user's attributes]}
\end{tcolorbox}
% \vspace{-4mm}
% \caption{Prompt for dialogue generation/regeneration (Step 1+).}
% \vspace{-4mm}
% \label{fig:prompt_his_gen_step1add}
% \end{figure}
%%%%%%%%%%%%Prompt for Dialogue Generation/Regeneration (Step 1+)%%%%%%%%%%%%%%%%%

%%%%%%%%%%%%Prompt for Dialogue Improvement%%%%%%%%%%%%%%%%%
% \begin{figure}[htbp]
% \centering
\begin{tcolorbox}[
  enhanced,
  % breakable,
  title={\small Dialogue Improvement},
  attach boxed title to top center={yshift=-3mm,yshifttext=-1mm},
  colback=white,
  colframe=black,
  colbacktitle=gray!80!red!70,
  coltitle=white,
  fonttitle=\bfseries,
  boxed title style={size=small,colframe=black},
  separator sign={},
  % width=0.8\textwidth,
]
\small
\texttt{Given the following dialogue:\\
\{User: Parks are usually crowded, and libraries can be too quiet. What else is there?\\
AI: You might enjoy a botanical garden or perhaps a quiet café with a nice view.\}\\
\\
Modify this dialogue to make the user message implicitly reflect the following user attribute:\\
\{agreeableness: low\}\\
Guidelines:\\
1. Do not explicitly mention the user attribute.\\
2. The user message should implicitly reflect the given attribute, allowing it to be inferred from the context.\\
3. Maintain a natural, single-round dialogue between a user and an AI assistant.\\
\\
Format the modified dialogue strictly as follows:\\
User: [Brief user message that implicitly reflects the given attribute]\\
AI: [Brief AI response that naturally responds to the user's message without referencing the user's attributes]}
\end{tcolorbox}
\vspace{20mm}
% \vspace{-4mm}
% \caption{Prompt for dialogue improvement.}
% \vspace{-4mm}
% \label{fig:prompt_his_improve}
% \end{figure}
%%%%%%%%%%%%Prompt for Dialogue Improvement%%%%%%%%%%%%%%%%%
~
%%%%%%%%%%%%Prompt for Attribute-Dialogue Alignment Examination%%%%%%%%%%%%%%%%%
% \begin{figure}[htbp]
% \centering
\begin{tcolorbox}[
  enhanced,
  % breakable,
  title={\parbox{0.8\linewidth}{\centering \small Attribute-Dialogue Alignment Examination}},
  attach boxed title to top center={yshift=-3mm,yshifttext=-1mm},
  colback=white,
  colframe=black,
  colbacktitle=gray!80!red!70,
  coltitle=white,
  fonttitle=\bfseries,
  boxed title style={size=small,colframe=black},
  separator sign={},
  % width=0.8\textwidth,
]
\small
\texttt{Examine the following dialogue:\\
\{User: Parks are usually crowded, and libraries can be too quiet. What else is there?\\
AI: You might enjoy a botanical garden or perhaps a quiet café with a nice view.\}\\
\\
Can you implicitly infer the user's attribute: \{agreeableness: low\} from the dialogue?\
"Implicitly" means the attribute can be reasonably inferred without direct mention, based on the user's language, concerns, or behavior in the dialogue.\\
Answer (yes/no): [Your answer here]}
\end{tcolorbox}
% \vspace{-4mm}
% \caption{Prompt for attribute-dialogue alignment examination.}
% \vspace{-4mm}
% \label{fig:prompt_his_exam}
% \end{figure}
%%%%%%%%%%%%Prompt for Attribute-Dialogue Alignment Examination%%%%%%%%%%%%%%%%%

%%%%%%%%%%%%Prompt for Dialogue Consistency Check%%%%%%%%%%%%%%%%%
% \begin{figure}[htbp]
% \centering
\begin{tcolorbox}[
  enhanced,
  % breakable,
  title={\small Dialogue Consistency Check},
  attach boxed title to top center={yshift=-3mm,yshifttext=-1mm},
  colback=white,
  colframe=black,
  colbacktitle=gray!80!red!70,
  coltitle=white,
  fonttitle=\bfseries,
  boxed title style={size=small,colframe=black},
  separator sign={},
  % width=0.8\textwidth,
]
\small
\texttt{Examine the following dialogue:\\
\{User: Everywhere is full of people, and libraries are just boring. What's left that's even worth bothering with?\\
AI: You might enjoy a botanical garden or perhaps a quiet café with a nice view.\}\\
\\
Is this dialogue consistent with the following user attribute(s)?\\
\{\{'extraversion': 'low', 'agreeableness': 'low', 'residence': 'rural', 'health': 'healthy'\}\}\\
Consistency definition: The dialogue content does not contradict any of the listed user attribute(s).\\
Answer yes or no only.\\
Answer (yes/no): }
\end{tcolorbox}
% \vspace{-4mm}
% \caption{Prompt for dialogue consistency check.}
% \vspace{-4mm}
% \label{fig:prompt_his_consistency}
% \end{figure}
%%%%%%%%%%%%Prompt for Dialogue Consistency Check%%%%%%%%%%%%%%%%%

\newpage
%%%%%%%%%%%%%%%%%%%%%%%%%%%%%%%%%%%%%%%%%%%%%%%%%%%%%%%%%%%%%%%%%%%%%%%%%%%%%%%%%%%%%%%%%%

%%%%%%%%%%%%%%%%%%%%%%%%%%%%%%%%%%%%%%%%%%%%%%%%%%%%%%%%%%%%%%%%%%%%%%%%%%%%%%%%%%%%%%%%%%%%%%%%%%%%%%%%%%%%%%%%%%%
% Construction Details
%%%%%%%%%%%%%%%%%%%%%%%%%%%%%%%%%%%%%%%%%%%%%%%%%%%%%%%%%%%%%%%%%%%%%%%%%%%%%%%%%%%%%%%%%%%%%%%%%%%%%%%%%%%%%%%%%%%

\section{Evaluation Details}
\addcontentsline{loa}{section}{\numberline{D}Evaluation Details}
\subsection{Evaluation Standard for \gptf-Score}
\label{app:gpt4o_score}
\addcontentsline{loa}{subsection}{\numberline{D.1}Evaluation Standard for \gptf-Score}
We define the evaluation standard for \gptf-Score with 4 criteria, illustrated by prompt below. As this prompt serves as a formal scoring template, we use parameter names in \{\} rather than specific examples for a cleaner presentation of the evaluation criteria. Due to evaluation costs, we randomly sample 10 items from each 100-item task for \gptf-Score evaluation.

%%%%%%%%%%%%Evaluation Standard for \gptf-Score%%%%%%%%%%%%%%%%%
% \begin{figure}[htbp]
% \centering
\begin{tcolorbox}[
  enhanced,
  % breakable,
  title={\small Evaluation Standard for \gptf-Score},
  attach boxed title to top center={yshift=-3mm,yshifttext=-1mm},
  colback=gray!5, 
  colframe=gray!80, 
  colbacktitle=gray!50!blue!30,  
  coltitle=white,
  fonttitle=\bfseries,
  boxed title style={size=small,colframe=gray!80},
  separator sign={},
  % width=0.8\textwidth,
]
\small
\texttt{Suppose you are a user with the following attributes: \{\phcolor{gt\_related\_attributes}\}. Your request is: \{\phcolor{user\_question}\}. Now evaluate the following analysis and answers to your request:\\
Agent Analysis: \{\phcolor{predict\_analysis}\}\\
Agent Answer: \{\phcolor{predict\_answer}\}\\
\\
Evaluation Standard:\\
Conciseness (0-1 points): Answers should be concise and to the point.\\
Personalization (0-4 points): Does the analysis and answer specifically address the user's request and attributes?\\
Analysis Quality (0-4 points): Is the analysis logical, coherent, and well-reasoned? Here's an example of a good analysis: \{\phcolor{gt\_analysis}\}. \\
Answer Accuracy (0-5 points): Does the answer correctly address the request? Ground Truth Answer: \{\phcolor{gt\_answer}\}.\\
Total Score: Sum of all categories (0-14 points)\\
\\
Output your score in the following format:\\
{}[Conciseness: score, Personalization: score, Analysis Quality: score, Answer Accuracy: score]\\
{}[Total: score]\\
\\
For example:\\
{}[Conciseness: 1, Personalization: 4, Analysis Quality: 4, Answer Accuracy: 5]\\
{}[Total: 14]\\
\\
Your Output:}
\end{tcolorbox}
% \vspace{-4mm}
% \caption{Evaluation standard for \gptf-Score. The standard is presented as a prompt template, where content in \{\} denotes parameters to be filled with input values.}
% \vspace{-4mm}
% \label{fig:standard_gptf}
% \end{figure}
%%%%%%%%%%%%Evaluation Standard for \gptf-Score%%%%%%%%%%%%%%%%%

\subsection{Prompts for Five Reasoning Pathways}
\addcontentsline{loa}{subsection}{\numberline{D.2}Prompts for Five Reasoning Pathways}
In this part, we present the detailed prompts used for each reasoning pathway described in Section \ref{sec:reason_pathway}. For all pathways, we provide the model with the user dialogue history $H$ and current question $Q$. We show the 5 designed prompts (\basic, \cotA, \cotAs, \cotAAs, \cotTsAs) below.
Similar to the prompt illustration of evaluation standard for GPT-4o-Score in Appendix \ref{app:gpt4o_score}, we use placeholders (e.g., \{\phcolor{task}\}, \{\phcolor{user\_history}\}, \{\phcolor{user\_question}\}) rather than specific examples in these prompts for a clearer presentation of the differences between the five reasoning pathways. The \phcolor{attribute\_dict} stores the attributes from Table \ref{tab:attribute_table} in dictionary format.

%%%%%%%%%%%%Reasoning Prompt - \basic%%%%%%%%%%%%%%%%%
% \begin{figure}[htbp]
% \centering
\begin{tcolorbox}[
  enhanced,
  % breakable,
  title= {\small Reasoning Prompt - \basic},
  attach boxed title to top center={yshift=-3mm,yshifttext=-1mm},
  colback=gray!5,  
  colframe=gray!80, 
  colbacktitle=green!20!gray!40,
  coltitle=white,
  fonttitle=\bfseries,
  boxed title style={size=small,colframe=gray!80},
  separator sign={},
  % width=0.8\textwidth,
]
\small
\texttt{You are a helpful agent for the task: \{\phcolor{task}\}. The task requires the agent to \{\phcolor{task\_definition}\}.\\
You will leverage my history dialogue to respond to my request in a personalized way. My history dialogue is:\\
\{\phcolor{user\_history}\}\\
---Important Note---\\
- \{\phcolor{specific\_note\_for\_task}\}\\
- First, provide the reasoning process for your answer in Analysis: [..]. Then, present your formatted answer in Answer: [..].\\
- Output your response in the format below, do not omit the [] in your response:\\
Output: \\
- Analysis: [..]\\
- \{\phcolor{extra\_answer\_format\_for\_task}\}\\
My Request: \{\phcolor{user\_question}\}\\
Output: }
\end{tcolorbox}
% \vspace{-4mm}
% \caption{Reasoning prompt template - \basic.}
% \vspace{-4mm}
% \label{fig:prompt_basic}
% \end{figure}
%%%%%%%%%%%%Reasoning Prompt - \basic%%%%%%%%%%%%%%%%%

%%%%%%%%%%%%Reasoning Prompt - \cotA%%%%%%%%%%%%%%%%%
% \begin{figure}[htbp]
% \centering
\begin{tcolorbox}[
  enhanced,
  % breakable,
  title= {\small Reasoning Prompt - \cotA},
  attach boxed title to top center={yshift=-3mm,yshifttext=-1mm},
  colback=gray!5,  
  colframe=gray!80, 
  colbacktitle=green!20!gray!40,
  coltitle=white,
  fonttitle=\bfseries,
  boxed title style={size=small,colframe=gray!80},
  separator sign={},
  % width=0.8\textwidth,
]
\small
\texttt{You are a helpful agent for the task: \{\phcolor{task}\}. The task requires the agent to \{\phcolor{task\_definition}\}.\\
You will leverage my history dialogue and my inferred attributes to respond to my request in a personalized way. My history dialogue is:\\
\{\phcolor{user\_history}\}\\
You can infer my attributes from the user attribute group: \{\phcolor{attribute\_dict}\}\\
---Important Note---\\
- \{\phcolor{specific\_note\_for\_task}\}\\
- First, infer my attribute value of ALL the attributes in the user attribute group in Full Attribute Values: [...]. Next, provide the reasoning process for your answer in Analysis: [..]. Finally, present your formatted answer in Answer: [..].\\
- ALL the attribute values should be selected from the given user attribute group.\\
- Your reasoning should be concise and clear.\\
- Output your response in the format below, do not omit the [] in your response:\\
Output: \\
- Full Attribute Values: [attribute1: value1, attribute2: value2, ..., attribute12: value12], such as [age: child, income\_level: low\_income, ..., hobby: sports]\\
- Analysis: [..]\\
- \{\phcolor{extra\_answer\_format\_for\_task}\}\\
My Request: \{\phcolor{user\_question}\}\\
Output: }
\end{tcolorbox}
% \vspace{-4mm}
% \caption{Reasoning prompt template - \cotA.}
% \vspace{-4mm}
% \label{fig:prompt_CoT_A}
% \end{figure}
%%%%%%%%%%%%Reasoning Prompt - \cotA%%%%%%%%%%%%%%%%%

%%%%%%%%%%%%Reasoning Prompt - \cotAs%%%%%%%%%%%%%%%%%
% \begin{figure}[htbp]
% \centering
\begin{tcolorbox}[
  enhanced,
  % breakable,
  title={\small Reasoning Prompt - \cotAs},
  attach boxed title to top center={yshift=-3mm,yshifttext=-1mm},
  colback=gray!5,  
  colframe=gray!80, 
  colbacktitle=green!20!gray!40,
  coltitle=white,
  fonttitle=\bfseries,
  boxed title style={size=small,colframe=gray!80},
  separator sign={},
  % width=0.8\textwidth,
]
\small
\texttt{You are a helpful agent for the task: \{\phcolor{task}\}. The task requires the agent to \{\phcolor{task\_definition}\}.\\
You will leverage my history dialogue and my inferred attributes to respond to my request in a personalized way. My history dialogue is:\\
\{\phcolor{user\_history}\}\\
You can infer my attributes from the user attribute group: \{\phcolor{attribute\_dict}\}\\
---Important Note---\\
- \{\phcolor{specific\_note\_for\_task}\}\\
- First, based on my history dialogue and my request, infer related attributes you will use to answer my request in Attributes: [..]. Then, provide the reasoning process for your answer in Analysis: [..]. Finally, present your formatted answer in Answer: [..].\\
- You should select the user attributes that are most relevant to the request. Use no more than 5 attributes. The attributes should be selected from the given user attribute group.\\
- Your reasoning should be concise and clear.\\
- Output your response in the format below, do not omit the [] in your response:\\
Output: \\
- Attributes: [attribute1: value1, attribute2: value2, ...], such as [age: child, profession: student]\\
- Analysis: [..]\\
- \{\phcolor{extra\_answer\_format\_for\_task}\}\\
My Request: \{\phcolor{user\_question}\}\\
Output: }
\end{tcolorbox}
% \vspace{-4mm}
% \caption{Reasoning prompt template - \cotAs.}
% \vspace{-4mm}
% \label{fig:prompt_CoT_As}
% \end{figure}
%%%%%%%%%%%%Reasoning Prompt - \cotAs%%%%%%%%%%%%%%%%%

%%%%%%%%%%%%Reasoning Prompt - \cotAAs%%%%%%%%%%%%%%%%%
% \begin{figure}[htbp]
% \centering
\begin{tcolorbox}[
  enhanced,
  % breakable,
  title={\small Reasoning Prompt - \cotAAs},
  attach boxed title to top center={yshift=-3mm,yshifttext=-1mm},
  colback=gray!5,  
  colframe=gray!80, 
  colbacktitle=green!20!gray!40,
  coltitle=white,
  fonttitle=\bfseries,
  boxed title style={size=small,colframe=gray!80},
  separator sign={},
  % width=0.8\textwidth,
]
\small
\texttt{You are a helpful agent for the task: \{\phcolor{task}\}. The task requires the agent to \{\phcolor{task\_definition}\}.\\
You will leverage my history dialogue and my inferred attributes to respond to my request in a personalized way. My history dialogue is:\\
\{\phcolor{user\_history}\}\\
You can infer my attributes from the user attribute group: \{\phcolor{attribute\_dict}\}\\
---Important Note---\\
- \{\phcolor{specific\_note\_for\_task}\}\\
- First, infer my attribute value of ALL the attributes in the user attribute group in Full Attribute Values: [...]. Then, extract related attributes you will use to answer my request, and put them in Related Attribute Values: [..]. Next, provide the reasoning process for your answer in Analysis: [..]. Finally, present your formatted answer in Answer: [..].\\
- You should extract the user attributes that are most relevant to the request. Use no more than 5 attributes. ALL the attribute values should be selected from the given user attribute group.\\
- Your reasoning should be concise and clear.\\
- Output your response in the format below, do not omit the [] in your response:\\
Output: \\
- Full Attribute Values: [attribute1: value1, attribute2: value2, ..., attribute12: value12], such as [age: child, income\_level: low\_income, ..., hobby: sports]\\
- Related Attribute Values: [related\_attribute1, related\_attribute2, ...], such as [age: child, profession: student]\\
- Analysis: [..]\\
- \{\phcolor{extra\_answer\_format\_for\_task}\}\\
My Request: \{\phcolor{user\_question}\}\\
Output: }
\end{tcolorbox}
% \vspace{-4mm}
% \caption{Reasoning prompt template - \cotAAs.}
% \vspace{-4mm}
% \label{fig:prompt_CoT_AAs}
% \end{figure}
%%%%%%%%%%%%Reasoning Prompt - \cotAAs%%%%%%%%%%%%%%%%%

%%%%%%%%%%%%Reasoning Prompt - \cotTsAs%%%%%%%%%%%%%%%%%
% \begin{figure}[htbp]
% \centering
\begin{tcolorbox}[
  enhanced,
  % breakable,
  title={\small Reasoning Prompt - \cotTsAs},
  attach boxed title to top center={yshift=-3mm,yshifttext=-1mm},
  colback=gray!5,  
  colframe=gray!80, 
  colbacktitle=green!20!gray!40,
  coltitle=white,
  fonttitle=\bfseries,
  boxed title style={size=small,colframe=gray!80},
  separator sign={},
  % width=0.8\textwidth,
]
\small
\texttt{You are a helpful agent for the task: \{\phcolor{task}\}. The task requires the agent to \{\phcolor{task\_definition}\}.\\
You will leverage my history dialogue and my inferred attributes to respond to my request in a personalized way. My history dialogue is:\\
\{\phcolor{user\_history}\}\\
You can infer my attributes from the user attribute group: \{\phcolor{attribute\_dict}\}\\
---Important Note---\\
- \{\phcolor{specific\_note\_for\_task}\}\\
- First, identify what attribute keys are most related to answer my request in Attribute Key: [...]. Then, based on my history dialogue, infer attribute values of your identified attribute, and write it in Attribute Values: [..]. Next, provide the reasoning process for your answer in Analysis: [..]. Finally, present your formatted answer in Answer: [..].\\
- You should select the user attributes that are most relevant to the request. Use no more than 5 attributes. The attributes should be selected from the given user attribute group.\\
- Your reasoning should be concise and clear.\\
- Output your response in the format below, do not omit the [] in your response:\\
Output: \\
- Attribute Key: [attribute1, attribute2, ...], such as [age, profession]\\
- Attribute Values: [attribute1: value1, attribute2: value2, ...], such as [age: child, profession: student]\\
- Analysis: [..]\\
- \{\phcolor{extra\_answer\_format\_for\_task}\}\\
My Request: \{\phcolor{user\_question}\}\\
Output: }
\end{tcolorbox}
% \vspace{-4mm}
% \caption{Reasoning prompt template - \cotTsAs.}
% \vspace{-4mm}
% \label{fig:prompt_CoT_TsAs}
% \end{figure}
%%%%%%%%%%%%Reasoning Prompt - \cotTsAs%%%%%%%%%%%%%%%%%
\newpage
%%%%%%%%%%%%%%%%%%%%%%%%%%%%%%%%%%%%%%%%%%%%%%%%%%%%%%%%%%%%
%%%%%%%%%%%%%%%%%%%%%%%%%%%%%%%%%%%%%%%%%%%%%%%%%%%%%%%%%%%%
\section{Experiment Details}
\addcontentsline{loa}{section}{\numberline{E}Experiment Details}
\subsection{Performance Evaluation}
\addcontentsline{loa}{subsection}{\numberline{E.1}Performance Evaluation}
We present more evaluations and findings in this section.
\paragraph{Attribute Performance.}
\textit{ATF} variance on tasks varies more slightly than \textit{ATF} variance on models. \gptomini~ranks the third, which aligns with its documented limitations, as these attribute-related capabilities heavily rely on world knowledge.

\paragraph{GPT-4o-Score.}
Most models maintain an average \gptf-Score above 7, demonstrating their basic capability in personalization tasks. 
The two Llama models often output invalid unmeaningful \textit{responses} with template-like patterns (e.g., "- Analysis: [..] - Answer: [..]"), where meaningful content is replaced with "..". Similar problem have been observed in several prior studies \citep{chen2024calm, chen2024clear}. 

\subsection{Influence of Different Reasoning Pathways}
\addcontentsline{loa}{subsection}{\numberline{E.2}Influence of Different Reasoning Pathways}
The min-max normalization we used to normalized \textit{task accuracy} and \textit{\gptf-Score} to the range [0,1] is 
\begin{equation*}
x'_i = \frac{x_i - \min(x)}{\max(x) - \min(x)},
\end{equation*}
where $x_i$ represents the score of a specific reasoning pathway, and $\min(x)$, $\max(x)$ are the minimum and maximum scores among all pathways under the same metric.

\begin{figure}[t]
\centering
\includegraphics[width=1\linewidth]{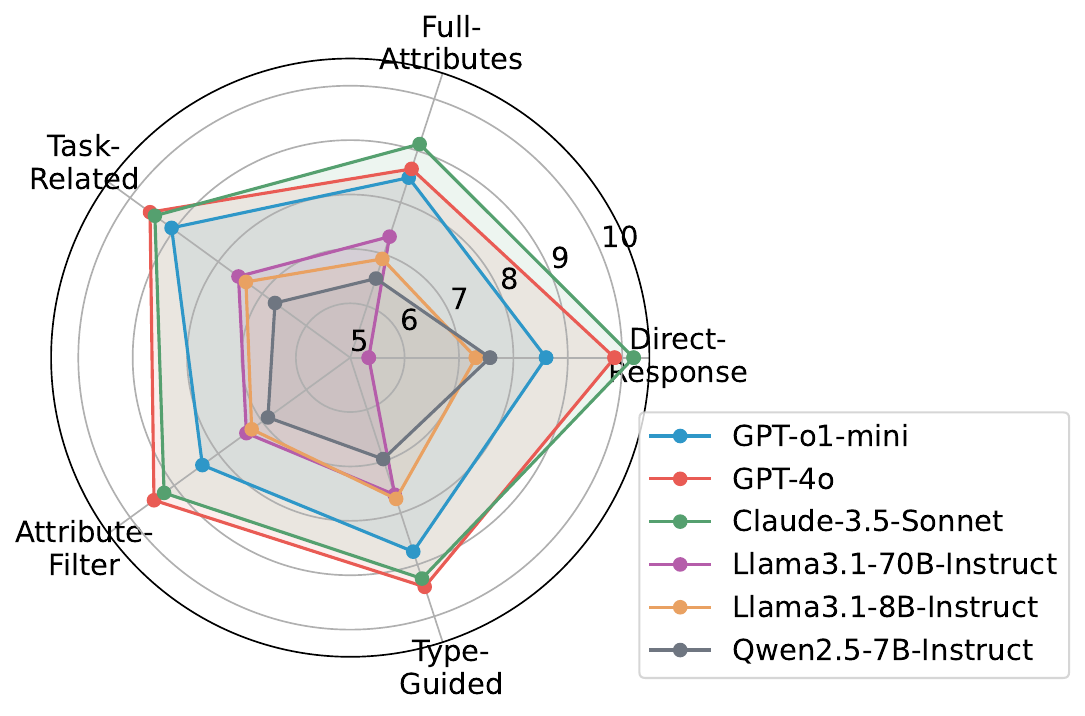}
\caption{Cross-model \textit{\gptf-Score}.}
\label{fig:Cross_model_comparison_gpt4o}
\end{figure}

We further provide the cross-model comparison on \textit{\gptf-Score} in Figure \ref{fig:Cross_model_comparison_gpt4o}. The discoveries are similar with those we report in Section \ref{sec:comp_reason_path}. Both \cotA~(inferring all possible attributes at the beginning) and \basic~(generating responses without attribute reasoning) show distinct performance patterns across different models. Models with stronger reasoning and information processing capabilities better adapt to these approaches, with some even achieving superior performance under certain metrics.
% \begin{figure}[t]
% \centering
% \includegraphics[width=1\linewidth]{Figures/Exp/CoT_model.pdf}
% \caption{Cross-model \textit{\gptf-Score}.}
% \label{fig:Cross_model_comparison_gpt4o}
% \end{figure}

\subsection{SFT on Trainset}
\addcontentsline{loa}{subsection}{\numberline{E.3}SFT on Trainset}
\label{app:task_acc_improve_cot}
\begin{table}[h]
    \centering
    \footnotesize
    \begin{tabular}{l|c}
        \toprule
        \textbf{Hyper-parameter} & \textbf{Value}\\
        \midrule
        Batch Size & 8\\
        Learning Rate & 2.0e-5 \\
        Number of Training Epochs & 3\\
        Warmup Ratio & 0.03\\
        \bottomrule
    \end{tabular}
    \caption{Hyper-parameters for SFT.}
    \label{tab:SFT_param}
\end{table}
We train \llamas~with 4 A100 GPU using LLaMA-Factory\footnote{\href{https://github.com/hiyouga/LLaMA-Factory}{https://github.com/hiyouga/LLaMA-Factory}}. Each of the 3 training processes takes 1-1.5 hours. Table \ref{tab:SFT_param} shows the hyper-parameters used in our SFT experiments. We use default values without tuning and report results for each evaluation experiment from a single run.

\begin{figure*}[t]
    \centering
    \includegraphics[width=1\linewidth]{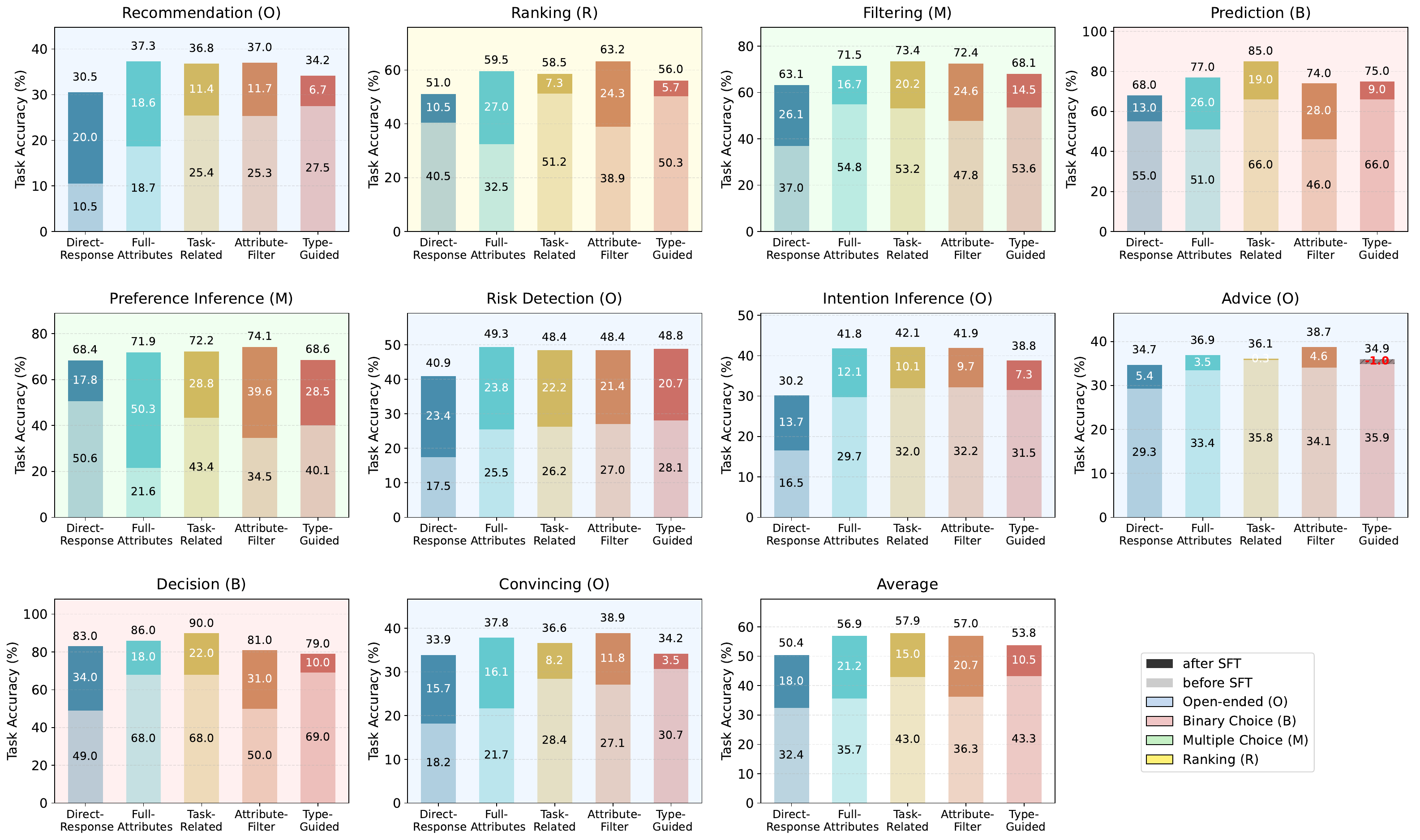}
    \caption{\textit{Task accuracy} improvement of \llamas~after SFT.}
    \label{fig:task_acc_improve_cot}
\end{figure*}

Figure \ref{fig:task_acc_improve_cot} illustrates the \textit{task accuracy} improvements achieved by \llamas~after SFT across various tasks and CoT prompts. Besides the conclusion in the main paper part, we find that: 
(1) The performance peak shifts from \cotTsAs~to \cotAs~-- an expected outcome given the training procedure utilizes the \cotAs~reasoning pathway.
(2) \cotTsAs, with its distinctly different reasoning pathway from \cotAs, underperformed compared to prompts that begin with $As$ or $A$.

%%%%%%%%%%%%%%%%%%%%%%%%%%%%%%%%%%%%%%%%%%%%%%%%%%%%%%%%%%%%%%%%%%%%%%%%%%%
\subsection{Automatic Quality Evaluation}
\label{app:auto_qual_eval}
\addcontentsline{loa}{subsection}{\numberline{E.4}Automatic Quality Evaluation}
\subsubsection{Baseline} 
For comparative analysis in diversity and fluency, we select several well-established open-ended QA datasets, including EmpatheticDialogues (ED) \citep{DBLP:conf/acl/RashkinSLB19}, Quora Question Pairs (QQP)\footnote{\href{https://quoradata.quora.com/First-Quora-Dataset-Release-Question-Pairs}{https://quoradata.quora.com/First-Quora-Dataset-Release-Question-Pairs}}, Natural Questions (NQ)\footnote{\href{https://ai.google.com/research/NaturalQuestions}{https://ai.google.com/research/NaturalQuestions}}, TriviaQA \citep{joshi-etal-2017-triviaqa}, Question Answering in Context (QuAC) \citep{choi-etal-2018-quac}, Conversational Question Answering (CoQA) \citep{reddy-etal-2019-coqa} and TopiOCQA \citep{adlakha2022topiocqa}.
\subsubsection{Diversity}
\label{app:auto_qual_eval_div}
To assess semantic diversity, we utilize NV-Embed-v2 \citep{lee2024nv}, a generalist embedding model, to compute embeddings for each question. We then calculate the average cosine similarity between all question pairs, where lower mean cosine similarity indicates greater semantic diversity.

For lexical diversity evaluation, we employ three length-insensitive metrics: Moving Average Type-Token Ratio (MATTR) \citep{covington2010mattr}, Measure of Textual Lexical Diversity (MTLD) \citep{mccarthy2005mtld}, and Hypergeometric Distribution Diversity (HD-D) \citep{mccarthy2010hd-d}. For meaningful comparisons across these metrics with different value ranges, we develop a unified metric called the \textbf{Lexical Diversity Score (LDS)}. The LDS formula, defined in Equation \eqref{eq:LDS}, normalizes these three metrics to a comparable scale through tangent transformation:
\begin{equation}
\label{eq:LDS}
\begin{split}
    \text{LDS} = &\left[\text{mtld} + \tan\left(\text{mattr}\cdot\frac{\pi}{2}\right)\right. \\&+ \left.\tan\left(\text{hdd}\cdot\frac{\pi}{2}\right)\right] / 3.
\end{split}
\end{equation}

The three metrics for lexical diversity evaluation -- Moving Average Type-Token Ratio (MATTR) \citep{covington2010mattr}, Measure of Textual Lexical Diversity (MTLD) \citep{mccarthy2005mtld}, and Hypergeometric Distribution Diversity (HD-D) \citep{mccarthy2010hd-d} -- are computed using the LexicalRichness package \citep{lex}.

Achieving leading performance in both semantic and lexical diversity in Figure \ref{fig:quality_auto}(a) demonstrates that \ipdialog~has broad coverage of diverse topics and contexts as well as rich vocabulary. 

\subsubsection{Fluency}
\label{app:auto_qual_eval_flu}
We evaluate fluency using perplexity scores computed by \llamas.
Perplexity, defined as the exponentiated average negative log-likelihood of a sequence, serves as a statistical measure of text fluency. A lower perplexity score indicates that the text follows more natural language patterns.

\subsubsection{Consistency}
\label{app:auto_qual_eval_const}
To evaluate dataset consistency, we randomly sample 1000 items from both the training and test sets (\ipdialog~benchmark) and generate multiple versions of ground truth answers: three from \gptf~(\gptf(1), \gptf(2), \gptf(3)) and one from \claude~(\claude(1)). 
We then evaluate six models on these samples: Llama3-8B-Instruct, Llama3-70B-Instruct, Qwen2.5-7B-Instruct, InternLM2.5-20B-Chat, Llama3.1-8B-Instruct, and Mistral-Nemo-Instruct. 
For each item, the models are provided with hidden user attributes and a user question and are asked to generate a response. 

\begin{figure}[t]
\begin{center}
%\framebox[4.0in]{$\;$}
\includegraphics[width=0.5\textwidth]{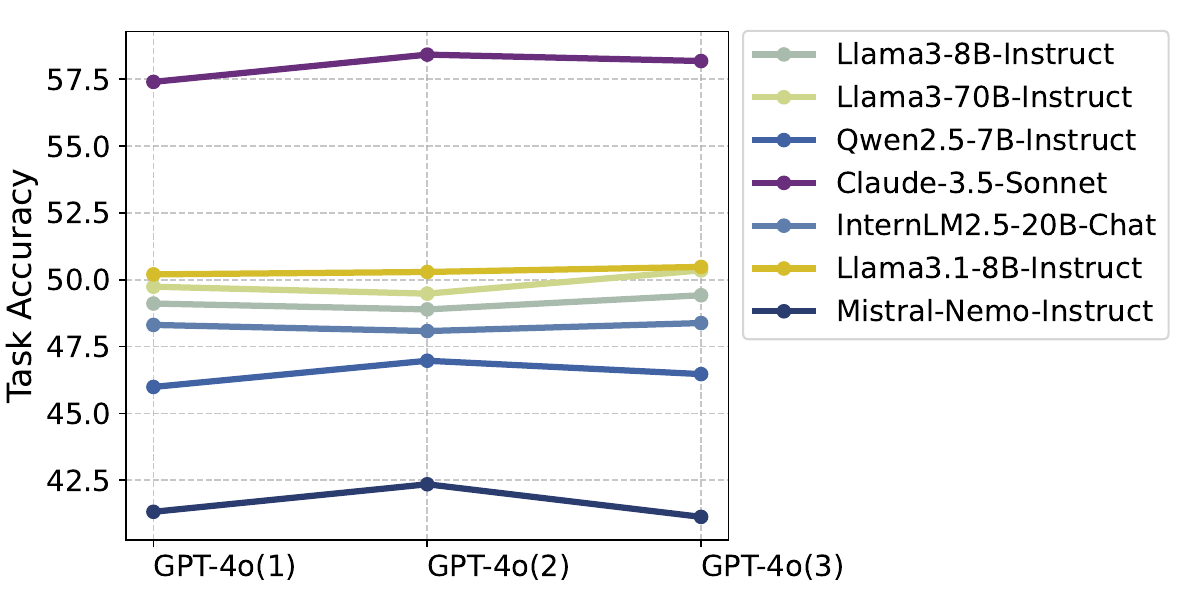}
\end{center}
\caption{Evaluation consistency check on answer versions of \gptf(1), \gptf(2), \gptf(3).}
\label{fig:consistency_3_gpt4o}
\end{figure}

In addition to the consistency analysis results in Figure \ref{fig:quality_auto}(b), we also provide an analysis of model performances including \claude~on 3 ground truth answer versions in Figure \ref{fig:consistency_3_gpt4o}, which proves the reliability of our dataset's ground truth answer.

%%%%%%%%%%%%%%%%%%%%%%%%%%%%%%%%%%%%%%%%%%%
\subsection{Human Study}
\addcontentsline{loa}{subsection}{\numberline{E.5}Human Study}
\label{app:human_study}
In this section, we present detailed information about our human evaluation study, which focused on the average human performance and quality of the generated data. We recruited annotators with diverse backgrounds to conduct the evaluation. All annotators were English-proficient and had at least a bachelor's degree, ensuring both demographic diversity and academic qualification in our participant pool. The annotators received fair compensation for their work, with all payments funded through our research group. The summarized annotator setting and averaged time used are shown in Table \ref{tab:ann_time}.

\begin{table*}[t]
\centering
\footnotesize
% \resizebox{0.4\textwidth}{!}{
\begin{tabular}{ccccm{2.5cm}m{3cm}}
\toprule
Experiment & Samples & Annotator & Ann. per sample & Annotation time per sample & Total annotation time per annotator\\ \midrule
Attribute Inference      &  100 &  2  & 1                       & \multicolumn{1}{c}{2.5min}          & \multicolumn{1}{c}{125min}\\
Task Accuracy     & 50  &  4  & 2                       & \multicolumn{1}{c}{3min}          & \multicolumn{1}{c}{75min}\\
\midrule
Fidelity     & 200  &  6  & 3                       & \multicolumn{1}{c}{15sec}          & \multicolumn{1}{c}{25min}\\
Attribute-Dialogue Align     & 200  &  4  & 2                       & \multicolumn{1}{c}{2min}          & \multicolumn{1}{c}{200min}\\
Attribute-Response Align     & 200  &  4  & 2                       & \multicolumn{1}{c}{2min}          & \multicolumn{1}{c}{200min}\\
\bottomrule
\end{tabular}
% }
\caption{Human study setting and annotation time.}
\label{tab:ann_time}
\end{table*}

\subsubsection{Human Performance}
\paragraph{Attribute Inference Accuracy.}
\label{app:attribute_inference}

\begin{table*}[t]
\centering
\footnotesize
% \resizebox{0.4\textwidth}{!}{
\begin{tabular}{cccccc}
\toprule
Method & Human & GPT-4o & Claude-3.5-Sonnet & Llama-3.1-70B & Random\\ \midrule
Precision($T_f$)      &  73.2 &  75.1  & \textbf{80.7}                       & 49.5          & 8.3\\
Precision($A_f$)      & 62.0  &  65.6  & \textbf{71.2}                       & 38.8          & 2.0\\
\bottomrule
\end{tabular}
% }
\caption{The performances of different methods or models on attribute inference test.}
\label{tab:attribute_inference}
\end{table*}
To evaluate human performance on attribute inference, we conduct a human study on 100 randomly sampled instances from \ipdialog~. Annotators are tasked with inferring attribute types and values from each of the historical dialogues. The annotation is performed using a predefined set of possible attributes outlined in Table \ref{tab:attribute_table}. 
The quality of annotations is assessed using two key metrics: Precision($T_f$) and Precision($A_f$), where $T_f$ represents the correctly identified attribute types and $A_f$ denotes the accurately predicted attribute values. For comparison, we also evaluate the performance of several advanced language models (GPT-4o, Claude-3.5-Sonnet, and Llama-3.1-70B) under the same experimental settings.
We divided randomly sampled 100 instances into two groups, each independently labeled by one annotator. Recognizing that the annotators might not possess prior knowledge of some attribute types, such as those from the Big Five personality traits, we provided detailed explanations of each attribute to ensure fair evaluation. The final evaluation result is shown in Table \ref{tab:attribute_inference}. In this experiment, \gptf~and \claude~outperform humans, successfully detecting subtle conversational cues that reflect user attributes.
Such capability requires advanced reading comprehension and extensive world knowledge.

\paragraph{Task Accuracy.}
\label{app:human_task_acc}
We randomly sampled 50 questions from five distinct close-ended task types: \textbf{ranking}, \textbf{filtering}, \textbf{prediction}, \textbf{preference inference}, and \textbf{decision}, with 10 instances from each category. These samples were then divided into two groups, with each group independently processed by two reviewers.
To assess human performance on these tasks, annotators answered questions based on a set of provided \textit{candidate attributes}. The human annotators achieved an average \textit{task accuracy} of 68.8, comparable to Llama3-70B-Instruct (68.6) but lower than \gptf~(81.8) and \claude~(76.4).

\subsubsection{Quality Analysis.}
\paragraph{Fidelity}
\label{app:fidelity_analysis}
We conduct a Turing test to evaluate whether human annotators could distinguish between AI-generated and human-produced utterances. 
Our evaluation corpus comprised 100 real dialogues and 100 synthetic dialogues. The real dialogues were sampled from the DailyDialog corpus \citep{li-etal-2017-dailydialog}, which is well-known for its diverse conversational topics and linguistic nuances. The synthetic dialogues were extracted from the user history dialogue of our \ipdialog~dataset. Both sets of dialogues were randomly sampled. To minimize length-related bias, we restricted each dialogue to contain between 25 and 35 tokens, thereby eliminating potential confounding factors that might affect participants' judgments. The average token count was comparable between the two sets: 29.17 for real dialogues and 29.52 for synthetic dialogues.
To maintain objectivity, dialogues were presented to participants in random order, and the source of each dialogue (real or synthetic) was not disclosed. Each dialogue was evaluated by three annotators from a pool of six participants. 

As shown in Table \ref{tab:fidelity_analysis}, participants achieve an accuracy rate of 52.2\%, only marginally outperforming random choice. The result indicates that our AI-generated dialogues are nearly indistinguishable from human-generated ones.
The inter-annotator agreement, measured by \textit{Fleiss' Kappa}, was 0.015. This value, being close to zero, indicates minimal consensus among annotators in distinguishing between human and AI-generated content. Such low agreement suggests that our synthetic dialogues achieved a level of naturalness comparable to human-generated ones.

\begin{table}[t]
\centering
\footnotesize
% \resizebox{0.4\textwidth}{!}{
\begin{tabular}{ccc}
\toprule
\diagbox{Pred.}{Label} & Human & AI  \\ \midrule
Human    &  201   &  187 \\
AI    &  187  &   112 \\
\bottomrule
\end{tabular}
% }
\caption{Fidelity analysis: distribution of predictions and true labels in the human-AI utterance classification.}
\label{tab:fidelity_analysis}
\end{table}

\paragraph{Attribute-dialogue Alignment.}
For the manual evaluation of attribute-dialogue alignment, we randomly sampled 200 instances for review, with each instance assessed by two evaluators. Four evaluators are involved in this experiment. The evaluators assessed whether the dialogue content provided adequate information for attribute inference. They were instructed to flag any instances where attributes could not be reliably inferred and provide brief explanations for these judgments.

92.0\% of the utterances are reviewed as accurately reflecting their corresponding ground truth attributes.
While our results demonstrate strong overall attribute-dialogue alignment, evaluators identified certain cases where they thought inferring user attributes was too arbitrary. For example:
\begin{itemize}
    \item In the utterance "\textit{Thanks! My girlfriends keep raving about Notion. Do you know if it has templates for studying or assignment tracking?}", evaluators questioned whether the use of "girlfriends" sufficiently indicates a female speaker. This hesitation is reasonable, yet sociolinguistic research provides supporting evidence: female speakers statistically use "girlfriends" more frequently than males when referring to female friends~\citep{articledns,quora_girlfriends}, whereas male speakers typically avoid this term due to its potential romantic connotation, opting instead for "female friends" or simply "friends." Furthermore, social network studies have shown that people typically maintain friendship circles dominated by their own gender~\citep{pew_friendship_2023,mjaavatn2016adolescents}. When someone casually mentions their 'girlfriends' in everyday conversation, it suggests they regularly interact with a female social group. Since people tend to socialize within same-gender circles, this pattern possibly indicates that the speaker is female.
    \item Similarly, in "\textit{As we review, I can't help but think of this checklist as the script for a blockbuster movie. Every detail needs to be in place for the perfect ending!}", evaluators questioned whether using movie metaphors indicates film interest. This critical perspective exemplifies thorough evaluation. However, the statement contains multiple film-specific elements: the person naturally uses industry terminology ("script," "blockbuster"), applies film production concepts to everyday tasks, and references narrative structure ("perfect ending"). When people repeatedly draw metaphors from a specific domain, it typically reflects their familiarity with and interest in that domain. Just as sports enthusiasts often use sports metaphors or musicians use musical analogies, this natural incorporation of film elements suggests some level of engagement with film media.
\end{itemize}
These examples demonstrate that cases seemingly too ambiguous for attribute inference may contain reasonable linguistic indicators for prediction. In everyday communication, humans also make probabilistic inferences about others based on subtle clues. Our dataset captures this inherent characteristic of human interaction, recognizing both its values and limitations. The evaluators' feedback highlights an important research direction: determining what linguistic patterns constitute sufficient evidence for attribute inference. This is crucial for developing AI systems that understand users naturally and respectfully.

\paragraph{Attribute-response Alignment.}
We measure the degree to which the analysis and responses align with the inferred attributes. 
With ground truth responses provided, annotators reviewed each GT instance in the dataset for attribute-response alignment. Four annotators are involved to examine 200 samples, with each sample examined by two annotators. They check each user question, its \textit{related attributes}, its ground truth analysis, and answers. 
The assessment used three key dimensions:
\begin{itemize}
\item \textbf{Attribute Consistency}: Whether the response properly incorporates and addresses all relevant attributes identified in the analysis phase.
\item \textbf{Analytical Coherence}: The logical flow between the attribute analysis and the final response.
\item \textbf{Analysis-Response Consistency}: Whether key insights from the analysis are properly reflected in the final response.
\end{itemize}
The review process revealed that 91.9\% of the evaluated samples demonstrated satisfactory alignment across all assessment criteria. We investigated the sample of errors and found that these instances frequently exhibited inconsistencies between the analysis and the final answer. Specifically, elements emphasized in the analysis are often not given corresponding importance in the final response. This misalignment suggests potential gaps in the translation of analytical insights into actionable components within the answers.

\section{AI Assistants In Research Or Writing}
\addcontentsline{loa}{section}{\numberline{F}AI Assistants In Research Or Writing}
This research was conducted with the assistance of AI tools for function documentation lookup during coding and grammar checking during the writing process.
\section{Case Study}
\addcontentsline{loa}{section}{\numberline{G}Case Study}
We present five examples from our dataset shown below. Note that the red texts, which provide additional explanations to enhance understanding, do not exist in the original dataset. The blue texts highlight the key content reflecting the \textit{related attribute}.

%%%%%%%%%%%%%%%%%%%%%%%%Case Study: Recommendation%%%%%%%%%%%%%%%%%%%%%
\begin{figure*}[htb]
\centering
\begin{tcolorbox}[
    enhanced,
    title=Case Study: Recommendation,
    boxed title style={
        size=small,
        minipage=6cm,
        center title,                  
        top=3mm,           
        bottom=3mm         
    },
    colback=gray!5,        
    colframe=black,       
    colbacktitle=yellow!60!green!50!black!45,   
    coltitle=gray!5!black!90,
    fonttitle=\bfseries,
    separator sign={},
    halign title=center,              
    toptitle=2pt,                                    
    bottomtitle=2pt                                  
]
\small
\textbf{[History]}

User: Oh no, I \textcolor{lightblue}{completely forgot} about my friend's birthday party tonight! Do you think I can grab a last-minute gift on the way there? \textcolor{darkred}{[Forgetting important things can be associated with lower conscientiousness.]}

AI: Absolutely! How about a nice gift card or a box of chocolates from a nearby store? You can still make it a great surprise.

User: Great idea! I'm excited to swoop in with something sweet. Plus, it'll be \textcolor{lightblue}{a chance to chat with everyone}. Do you think I should pick up a card too, or just chocolates? \textcolor{darkred}{[People with high extraversion tend to be more willing to chat with others.]}

AI: A card would be a thoughtful touch, and it gives you a chance to write a fun, personal message. It’ll definitely add to the celebration vibe!

User: Awesome, I'll \textcolor{lightblue}{swing by the local shop} to \textcolor{lightblue}{stay within budget}. Thanks for helping me make it special without going overboard! \textcolor{darkred}{[The user's shopping behavior - being able to shop but showing spending restraint - suggests a moderate income level.]}

AI: Sounds like a perfect plan! Enjoy the party, and I'm sure your friend will appreciate the thoughtful gesture.

User: I might have to \textcolor{lightblue}{dodge some traffic getting to a shop}, but I'm glad there are plenty of options nearby. \textcolor{darkred}{[The presence of traffic congestion in this area suggests an urban rather than rural location.]}

AI: With so many stores around, you're bound to find everything you need quickly. Safe travels, and enjoy the party!

\textbf{[Task]}

Recommendation: recommend products based on the needs of a user's specific attribute.

\textbf{[Question]}

I'm thinking about getting a car that's suitable for city life. What would you suggest?

\textbf{[Related Attributes]}

Conscientiousness: low,
extraversion: high,
income level: middle income,
residence: urban.

\textbf{[Analysis]}

Given the user's income level, \textcolor{lightblue}{practical and budget-friendly} options are recommended. An urban residence suggests a need for a \textcolor{lightblue}{compact, fuel-efficient} car that's \textcolor{lightblue}{easy to maneuver and park in city environments}. High extraversion might indicate an interest in a car \textcolor{lightblue}{with a stylish design and features} that accommodate an active lifestyle. Low conscientiousness suggests a preference for a car that \textcolor{lightblue}{requires low maintenance}.

\textbf{[Answer]}

Honda Fit, Toyota Yaris, Ford Fiesta. \textcolor{darkred}{[These are compact, fuel-efficient, stylish, and low-maintenance car models that are easy to park and suit an active, urban lifestyle on a budget.]}
\end{tcolorbox}
\vspace{-4mm}
\caption{Case study: Recommendation.}
\vspace{-4mm}
\label{fig:case_study_recommendation}
\end{figure*}
%%%%%%%%%%%%%%%%%%%%%%%%Case Study: Recommendation%%%%%%%%%%%%%%%%%%%%%

%%%%%%%%%%%%%%%%%%%%%%%%Case Study: Ranking%%%%%%%%%%%%%%%%%%%%%
\begin{figure*}[htbp]
\centering
\begin{tcolorbox}[
    enhanced,
    title=Case Study: Ranking,
    boxed title style={
        size=small,
        minipage=6cm,
        center title,                  
        top=3mm,          
        bottom=3mm        
    },
    colback=gray!5,        
    colframe=black,       
    colbacktitle=yellow!60!green!50!black!45,   
    coltitle=gray!5!black!90,
    fonttitle=\bfseries,
    separator sign={},
    halign title=center,              
    toptitle=2pt,                                    
    bottomtitle=2pt                                 
]
\small
\textbf{[History]}

User: I found this perfect spot with \BLUE{amazing lighting} yesterday that I can't wait to visit again soon. Any tips on \BLUE{tracking the best times for natural light} next time I go there? \RED{[Focusing on ideal lighting conditions is a characteristic of photographers.]}

AI: You might want to try using a sun-tracking app to check the position of the sun throughout the day. It can help find the ideal time for the kind of natural lighting you're looking for.

User: Are there any \BLUE{free} apps you recommend for sun tracking, so I can plan \BLUE{without spending extra}? \RED{[The emphasis on finding free apps to avoid additional costs suggests a need to minimize expenses, implying a lower income level.]}

AI: Definitely! There are several free apps available, like Sun Surveyor Lite and SunCalc, which should help you track the sun's path without any cost.

User: I might just \BLUE{stick} to visiting \BLUE{at the same time} as yesterday. It worked pretty well. \RED{[Preferring to return at the same time instead of experimenting with new options indicates a comfort with routine and familiarity, which suggests low openness.]}

AI: That sounds like a solid plan! Sometimes sticking with what works is the best approach. If you need any other tips, feel free to ask!

User: Maybe I'll capture some photos to show to our team during our next meeting. Visuals could really \BLUE{make our pitch more compelling} and highlight the presentation with all that natural light. \RED{[The emphasis on using visuals to "make our pitch more compelling" suggests the user may be involved in sales or business development roles where persuasive presentations are common.]}

AI: That sounds like a great idea! Sharing those photos can really help capture everyone's attention and set a positive mood. If you have any other thoughts or need assistance, let me know!

User: It’s amazing how \BLUE{urban landscapes} can create stunning backdrops too, especially with the right lighting. Any tips on capturing the uniqueness of \BLUE{city architecture} in photos? \RED{[The specific knowledge of urban lighting and architectural details suggests they likely reside in a city environment, where they can regularly observe these elements firsthand.]}

AI: Absolutely! Try focusing on unique angles or perspectives of the buildings, and use reflections in windows or puddles to add an interesting dimension. Playing with different times of day can also help capture the essence of urban life.

\textbf{[Task]}

Ranking: rank a series of given items according to the user's potential level of interest.

\textbf{[Question]}

I'm looking to prioritize the following economic topics in terms of how interesting they might be for me: 

A. inflation impacts on urban living B. trends in tech startups C. renewable energy investments D. financial literacy programs

\textbf{[Related Attributes]}

Hobby: photograph, income level: low income, openness: low, profession: sales worker, residence: urban.
% \end{tcolorbox}
% \newpage
% \begin{tcolorbox}[
%   enhanced,
%   breakable,
%   colback=white,
%   colframe=black,
%   separator sign={},
%   % width=0.8\textwidth,
% ]

\textbf{[Analysis]}

Given the user's low income level and urban residence, topics related to inflation impacts on urban living and financial literacy programs might be of higher interest as \BLUE{they could directly affect and benefit their personal situation}. The user's profession as a sales worker might \BLUE{create some interest in trends in tech startups}, as it can influence market dynamics and job opportunities. The user's low openness suggests a preference for familiar or immediately relevant topics, which would prioritize practical over abstract interests. Lastly, renewable energy investments, while broadly significant, might \BLUE{not be as immediately relevant or interesting to someone with this profile and hobby in photography}.

\textbf{[Answer]}

A, D, B, C

\end{tcolorbox}
\vspace{-4mm}
\caption{Case study: Ranking.}
\vspace{-4mm}
\label{fig:case_study_rank}
\end{figure*}
%%%%%%%%%%%%%%%%%%%%%%%%Case Study: Ranking%%%%%%%%%%%%%%%%%%%%%

%%%%%%%%%%%%%%%%%%%%%%%%Case Study: Filtering%%%%%%%%%%%%%%%%%%%%%
\begin{figure*}[htbp]
\centering
\begin{tcolorbox}[
    enhanced,
    title=Case Study: Filtering,
    boxed title style={
        size=small,
        minipage=6cm,
        center title,                  
        top=3mm,           
        bottom=3mm         
    },
    colback=gray!5,        
    colframe=black,       
    colbacktitle=yellow!60!green!50!black!45,   
    coltitle=gray!5!black!90,
    fonttitle=\bfseries,
    separator sign={},
    halign title=center,              
    toptitle=2pt,                                    
    bottomtitle=2pt                                  
]
\small
\textbf{[History]}

User: Hey AI, I just got back from \BLUE{hosting a dinner party}, and it was \BLUE{so much fun}! We played games, exchanged stories, and I think everyone \BLUE{had a great time}. Any suggestions for what we could do next week to \BLUE{keep the energy up}? \RED{[The speaker's enthusiasm for socializing, hosting, engaging in interactive activities, and planning future gatherings reflects high extraversion, as they derive energy from social interactions and enjoy lively group activities.]}

AI: That sounds amazing! How about organizing a themed karaoke night? You could even set up a small stage area to really amp up the excitement.

User: That sounds like a blast! I already have a speaker and a laptop, so I think I can make it work without \BLUE{breaking the bank}. Any tips on how to make it feel special \BLUE{without spending too much}? \RED{[The user's priority on cost-effective solutions over spending on extras suggests a low income level.]}

AI: Absolutely! You could create a fun playlist with everyone's favorite songs or decorate the space with some DIY lights and props for a festive touch. Maybe even make a karaoke contest out of it with some homemade certificates as prizes!

User: Great ideas! I can check with my neighbor about \BLUE{borrowing some hay bales} to use for seating, and maybe we can have it outside \BLUE{under the stars}. It's \BLUE{pretty quiet} around here, so we shouldn't bother anyone. \RED{[The user's mention of borrowing hay bales for seating, hosting an outdoor event under the stars, and describing the area as "pretty quiet" indicates a rural residence, where open space and agricultural resources are more accessible.]}

AI: That sounds charming! An outdoor setup will make it memorable, and you could even hang string lights between trees for a cozy atmosphere. Don't forget to have some warm drinks ready if it gets chilly!

\textbf{[Task]}

Filtering: filter a given list of items based on user attributes and preferences, retaining suitable content and removing irrelevant items.

\textbf{[Question]}

Help me filter these transportation options for tourists. Which ones should I consider? 

A. Private yacht rental B. Public subway system C. Guided city walking tour D. Shared ride-hailing service E. Camper van rental

\textbf{[Related Attributes]}

Extraversion: high, income level: low income, residence: rural.

\textbf{[Analysis]}

The user's low income level suggests \BLUE{cost-effective} transportation options are preferable. High extraversion indicates they may \BLUE{enjoy social environments}. Their rural residence might mean they appreciate opportunities to \BLUE{immerse in more urban or communal settings}. The public subway system and shared ride-hailing service both align with these considerations, offering affordable and potentially social travel experiences.

\textbf{[Answer]}

B, D
\end{tcolorbox}
\vspace{-4mm}
\caption{Case study: Filtering.}
\vspace{-4mm}
\label{fig:case_study_filter}
\end{figure*}
%%%%%%%%%%%%%%%%%%%%%%%%Case Study: Filtering%%%%%%%%%%%%%%%%%%%%%

%%%%%%%%%%%%%%%%%%%%%%%%Case Study: Prediction%%%%%%%%%%%%%%%%%%%%%
\begin{figure*}[htbp]
\centering
\begin{tcolorbox}[
    enhanced,
    title=Case Study: Prediction,
    boxed title style={
        size=small,
        minipage=6cm,
        center title,                 
        top=3mm,           
        bottom=3mm         
    },
    colback=gray!5,        
    colframe=black,        
    colbacktitle=yellow!60!green!50!black!45,   
    coltitle=gray!5!black!90,
    fonttitle=\bfseries,
    separator sign={},
    halign title=center,              
    toptitle=2pt,                                    
    bottomtitle=2pt                                  
]
\small
\textbf{[History]}

User: I \BLUE{finished organizing my schedule for the week}, \BLUE{making sure each task has a specific time slot}. \BLUE{Can you remind me} about my morning meeting tomorrow, just \BLUE{in case something changes}? \RED{[The proactive scheduling, attention to task timing, and desire for reminders reflect high conscientiousness, as they demonstrate thorough planning and a strong sense of responsibility.]}

AI: Of course, I'll remind you about your meeting and keep an eye out for any updates. Let me know if there's anything else you need help with!

User: I'll \BLUE{keep to myself} for now. \RED{[The choice to stay reserved and avoid social interaction indicates low extraversion.]}

AI: Alright, feel free to reach out when you want.

User: Actually, could you also check if any updates are needed for the itinerary of my business trip next week? I'm flying \BLUE{first class} and want to make sure everything is in order. \RED{[The ability to afford first-class airfare for a business trip suggests a high-income level.]}

AI: Certainly, I'll verify all the details of your trip itinerary and ensure that everything is up to date and smoothly arranged for you.

User: On second thought, I'd love some recommendations for things to do and places to explore around my destination. I have some free time during my trip and \BLUE{I'm curious to see what I might discover}! \RED{[The interest in exploring new activities and discovering unfamiliar places during the trip indicates high openness.]}

AI: Absolutely! I'll put together a list of suggestions for unique experiences and must-see spots at your destination. Let me know if you want to focus on anything specific!

User: I'm looking forward to some downtime and hoping to \BLUE{explore some serene spots} \BLUE{like the ones I'm used to back home}. Anything like that available? \RED{[The specific mention of finding serenity "like home" during downtime may indicate rural residence, as rural dwellers often seek familiar natural tranquility when visiting cities - a pattern less common among urban residents.]}

AI: Certainly, I'll find some serene nature trails and charming local farms at your destination for you to explore. You'll have a peaceful and rewarding experience for sure!

\textbf{[Task]}

Prediction: predict whether a user will take a specific action or make a particular decision.

\textbf{[Question]}

Will I opt for a carpool arrangement to reduce my travel expenses?

\textbf{[Related Attributes]}

Conscientiousness: high, extraversion: low, income level: high income, openness: high, residence: rural.

\textbf{[Analysis]}

While high income may reduce the need to cut travel costs, a high level of openness suggests you are open to new experiences like carpooling. High conscientiousness may drive you towards efficient and environmentally friendly choices. However, low extraversion might make you \BLUE{less inclined to share rides with others}. Living in a rural area may also \BLUE{influence the practicality and availability of carpooling}.

\textbf{[Answer]}

No.
\end{tcolorbox}
\vspace{-4mm}
\caption{Case study: Prediction.}
\vspace{-4mm}
\label{fig:case_study_predict}
\end{figure*}
%%%%%%%%%%%%%%%%%%%%%%%%Case Study: Prediction%%%%%%%%%%%%%%%%%%%%%

%%%%%%%%%%%%%%%%%%%%%%%%Case Study: Convincing%%%%%%%%%%%%%%%%%%%%%
\begin{figure*}[htbp]
\centering
\begin{tcolorbox}[
    enhanced,
    title=Case Study: Convincing,
    boxed title style={
        size=small,
        minipage=6cm,
        center title,                  
        top=3mm,           
        bottom=3mm         
    },
    colback=gray!5,        
    colframe=black,        
    colbacktitle=yellow!60!green!50!black!45,   
    coltitle=gray!5!black!90,
    fonttitle=\bfseries,
    separator sign={},
    halign title=center,              
    toptitle=2pt,                                    
    bottomtitle=2pt                                  
]
\small
\textbf{[History]}

User: I just finished organizing my tasks for the week. Could you \BLUE{remind me to review my plan every morning at 8 AM}? \RED{[The structured task organization and desire for daily 8 AM reviews indicates high conscientiousness through the disciplined planning habits of the user.]}

AI: Sure, I'll set a reminder for you every morning at 8 AM to review your weekly plan.

User: Thanks! I was thinking of adding \BLUE{"plan my next trip"} to the schedule. Do you think that's a good idea? \RED{The interest of the user in planing the next trip implies that travel is an activity they enjoy and prioritize.}

AI: That sounds like a great addition! It could be exciting to explore new destinations as part of your weekly routine.

User: I'm actually \BLUE{a bit worried} that I might forget something important during my trip planning. Do you think I should \BLUE{create a checklist to make sure I cover everything}? \RED{[The expression of worry about forgetting trip details and immediate request for organizational tools like checklists suggests heightened anxiety about potential oversights - a characteristic often associated with high neuroticism.]}

AI: Creating a checklist is always a smart move to ensure nothing gets overlooked. I can help you set one up if you'd like!

\textbf{[Task]}

Convincing: convince a user to do something.

\textbf{[Question]}

Encourage me to avoid single-use plastics during my travels.

\textbf{[Related Attributes]}

conscientiousness: high, hobby: travel, neuroticism: high.

\textbf{[Analysis]}

The user has high conscientiousness, suggesting they are organized and responsible, which \BLUE{aligns with the goal of reducing plastic waste.} Their interest in travel is relevant as it directly relates to the context of avoiding single-use plastics while traveling. Additionally, their high neuroticism might \BLUE{influence their sensitivity to environmental concerns or anxieties about contributing to pollution}, which could be addressed by taking positive actions like reducing plastic use.

\textbf{[Answer]}

As someone who is conscientious and passionate about travel, you have a unique opportunity to \BLUE{make a positive impact on the environments you explore}. By avoiding single-use plastics, you can \BLUE{help preserve the natural beauty of your travel destinations for future visits and generations}. Plus, taking this eco-friendly approach could ease any environmental concerns you may have, \BLUE{allowing you to enjoy your travels with peace of mind}, knowing you're contributing to a healthier planet.

\end{tcolorbox}
\vspace{-4mm}
\caption{Case study: Convincing.}
\vspace{-4mm}
\label{fig:case_study_convince}
\end{figure*}
%%%%%%%%%%%%%%%%%%%%%%%%Case Study: Convincing%%%%%%%%%%%%%%%%%%%%%

\end{document}